\documentclass[11pt]{article}

\usepackage{geometry}
\usepackage{setspace}
\usepackage{caption} 
\usepackage{graphicx}
\usepackage{enumitem}
\usepackage{xspace}
\usepackage{mdframed}
\usepackage{ifthen}
\usepackage{xcolor}
\usepackage{mathabx}
\usepackage{enumitem}
\usepackage{subcaption}

\usepackage{mathptmx}
\usepackage[T1]{fontenc}

\usepackage[natbibapa]{apacite}

\usepackage{multirow}

\newenvironment{biseabstract}{%
\begin{quote} \bf}
{\end{quote}}

\newenvironment{bisekeywords}{%
\begin{quote} \it \textbf{Keywords -}}
{\end{quote}}

\mdfdefinestyle{RQFrame}{
 outerlinewidth=0pt,
 skipabove=0pt,
 skipbelow=0pt,
 innertopmargin=6pt,
 innerbottommargin=0pt,
 linewidth=0pt,
 topline=false,
 rightline=false,
 leftline=false,
 innerrightmargin=4pt,
 innerleftmargin=4pt}

\newcounter{RQCounter}
\newcommand{\RQ}[2]{
\refstepcounter{RQCounter} \label{#1}
\begin{mdframed}[style=RQFrame]\noindent
	\textbf{RQ}$_{\arabic{RQCounter}}$.~\emph{#2}
\end{mdframed}
}
\newcommand{\hr}[1]{\textbf{RQ}$_{\ref{#1}}$}
\definecolor{Gray}{gray}{0.9}

\newboolean{showcomments}
\setboolean{showcomments}{true} 
\ifthenelse{\boolean{showcomments}}
  {\newcommand{\nb}[2]{
    \fcolorbox{Gray}{yellow}{\bfseries\sffamily\scriptsize#1}
    {\sf\small$\blacktriangleright$\textit{#2}$\blacktriangleleft$}
   }
   
  }
  {\newcommand{\nb}[2]{}
   
  }

\newcommand{\bpm}{BPM\xspace}
\newcommand{\bpmml}{BPM+ML\xspace}
\newcommand{\act}[1]{\texttt{#1}}
\newcommand{\feat}[1]{\textit{#1}}
\newcommand{\lab}[1]{\textsc{#1}}

\newcommand{\adela}{{BA1}\xspace}
\newcommand{\anti}{{BA2}\xspace}
\newcommand{\pierluigi}{{BB1}\xspace}
\newcommand{\claudio}{{BB2}\xspace}
\newcommand{\manuel}{{MA1}\xspace}
\newcommand{\han}{{MA2}\xspace}
\newcommand{\francesco}{{MB1}\xspace}
\newcommand{\niek}{{MB2}\xspace}

\newlist{question_answer}{itemize}{1}
\setlist[question_answer]{label=\textbf{Q:}}
\newcommand\itema{\item[\textbf{A:}]}

\title{Explainable Predictive Process Monitoring: A User Evaluation} 

\author
{
  Williams Rizzi$^{1,2\ast}$,
  Marco Comuzzi$^{4}$, 
  Chiara Di Francescomarino$^{1}$, 
  Chiara Ghidini$^{1}$, 
  Suhwan Lee$^{5}$, \\
  Fabrizio Maria Maggi$^{2}$, 
  Alexander Nolte$^{3,6}$ 
\\
\normalsize{$^{1}$Fondazione Bruno Kessler, Via Sommarive, 18, 38123 Povo, Italy}\\
\normalsize{$^{2}$Free University of Bozen-Bolzano, Piazza Università, 5, 39100 Bolzano, Italy}\\
\normalsize{$^{3}$University of Tartu, Narva maantee 18, 51009 Tartu, Estonia}\\
\normalsize{$^{4}$Ulsan National Institute of Science and Technology, Ulsan, South Korea}\\
\normalsize{$^{5}$Utrecht University, Utrecht, The Netherlands}\\
\normalsize{$^{6}$Carnegie Mellon University, Pittsburgh, PA, USA}\\
\\
\normalsize{$^\ast$To whom correspondence should be addressed; E-mail: wrizzi@fbk.eu}
}

\date{}

\begin{document} 

\baselineskip24pt

\maketitle 


\begin{biseabstract}
Explainability is motivated by the lack of transparency of black-box Machine Learning  approaches, which do not foster trust and acceptance of Machine Learning algorithms. This also happens in the Predictive Process Monitoring field, where predictions, obtained by applying Machine Learning techniques, need to be explained to users, so as to gain their trust and acceptance. 
In this work, we carry on a user evaluation on  explanation approaches for Predictive Process Monitoring  aiming at investigating whether and how the explanations provided (i) are understandable; (ii) are useful in decision making tasks; (iii) can be further improved for process analysts, with different Machine Learning expertise levels. {The results of the user evaluation show that, although explanation plots are overall understandable and useful for decision making tasks for Business Process Management users --- with and without experience in Machine Learning --- differences exist in the comprehension and usage of different plots, as well as in the way users with different Machine Learning expertise understand and use them.}

\end{biseabstract}

\begin{bisekeywords}
Predictive Process Monitoring, Process Mining, Explainable Artificial Intelligence, Explanation Plots, Qualitative Observational Study
\end{bisekeywords}



\section{Introduction}
\label{sec:intro}
Predictive Process Monitoring (PPM) is a branch of Process Mining that aims at providing predictions on the future of an ongoing process execution by leveraging past historical execution traces. An increasing number of PPM approaches leverage machine and deep learning techniques in order to learn from past historical execution traces the outcome of an ongoing process execution, the time remaining till the end of an ongoing execution, or the next activities that will be performed.

In many of these applications, users are asked to trust a model that supports them in making decisions. Therefore, understanding the rationale behind the predictions would certainly help users decide when to trust or not to trust these systems. In recent years, Explainable Artificial Intelligence (XAI) has been investigating the problem of explaining machine learning models so as to foster trust and acceptance of these models. Some of the recent XAI approaches have also been applied and investigated in the field of PPM~\citep{DBLP:conf/icpm/GalantiCLCN20,DBLP:conf/icsoc/Velmurugan0MS21,DBLP:conf/bpm/RizziFM20} in order to make a predictor returning, besides predictions, also prediction explanations (for instance in the form of \textit{explanation plots}).
None of these works has, however, investigated whether users actually understand and use these plots.

In this paper, we focus on investigating whether the users actually understand the explanation plots returned by XAI techniques in the context of PPM problems, whether these plots are able to support users when they need to make decisions, as well as how to improve them. Specifically, we aim at answering the following research questions:
\begin{enumerate}[label=]
\item \RQ{RQ1}{How do users make sense of explanation plots?}
\item \RQ{RQ2}{How can explanation plots support users in decision making?}
\item \RQ{RQ3}{How can explanation plots be improved?}
\end{enumerate}

In order to answer these research questions, we carried out a user evaluation with users working in the field of PPM and with different levels of Machine Learning (ML) expertise. We provided them with a problem, with some predictions and corresponding explanation plots. We hence asked the users (i) some questions for checking their comprehension of the plots, (ii) to make a decision by leveraging the plots, and (iii) how they would change the explanation plots to make them more useful and understandable. {The results of the user evaluation show that, although explanation plots are overall understandable and useful for decision making tasks for all participants, differences exist in the comprehension and usage of different plots, as well as in the way users with different ML expertise understand and use them. In particular, the study reveals that, while on the one side for ML experts understanding the explanation plots is easier than for participants without ML experience, on the other hand, using the plots for making decisions is easier for participants who do not have ML experience. ML experts, indeed, tend to be more conservative in their decisions, feeling that explanation plots, mainly showing correlations and not causalities, do not provide them with enough evidence to make specific recommendations. Moreover,  the evaluation carried out revealed interesting suggestions and desiderata for explanation plots, such as the need for interactive elements in the user interface as well as for what-if analysis support.}

In the next sections, we first provide some background knowledge useful for understanding the remainder of the paper (Section~\ref{sec:background}) and summarize the related work (Section~\ref{sec:relatedwork}). Then, Section~\ref{sec:method} describes the methodology used for carrying out the user evaluation and answering the research questions. Finally, Section~\ref{sec:findings} and Section~\ref{sec:disc} report and discuss the findings of the user evaluation.

\section{Background}
\label{sec:background}
In this section, we report the main concepts discussed in this paper, i.e., Predictive Process Monitoring (Section~\ref{sec:ppm}), Explainability approaches (Section~\ref{sec:explainability}), as well as Explainability applied to the Predictive Process Monitoring field (Section~\ref{sec:explainability_ppm}).
\subsection{Predictive Process Monitoring}
\label{sec:ppm} 
PPM~\citep{DBLP:conf/caise/MaggiFDG14, DBLP:reference/bdt/Francescomarino19} is a branch of process mining \citep{DBLP:conf/bpm/AalstAM11} that aims at exploiting event logs of past historical process execution traces to predict how ongoing (uncompleted) process executions will unfold up to their completion. 
Typical examples of predictions on the future of an execution trace relate to its completion time, to the fulfilment or violation of a certain predicate, or to the next sequence of activities that will be executed.

PPM approaches are typically characterized by two phases: a \textit{training phase}, in which a predictive model is learned from historical traces, and a \textit{prediction phase}, in which the predictive model is queried for predicting the future developments of an ongoing case.
Recent works addressing PPM challenges mainly leverage machine learning or statistical models, i.e., implicit models of the process rather than explicit process models~\citep{Maggi:CAiSE2014,Di-Francescomarino:2016aa,DBLP:conf/caise/TaxVRD17,DBLP:conf/bpm/Camargo19,Evermann2017}.

PPM approaches can be classified based on the predictions they provide~\citep{marquez-chamorro_predictive_2018,DBLP:conf/bpm/Francescomarino18}:
\begin{enumerate}[label=(\roman*)]
    \item \textit{numeric predictions}, i.e., continuous measures. Typical examples of numeric predictions are the remaining time of an ongoing execution, its duration or cost~\citep{Aalstetal:2011,Polatoetal:2018,vanDongen2008,Folino2013,DBLP:journals/tist/VerenichDRMT19}.
    \item \textit{outcome-based predictions}, i.e., predictions related to categorical or boolean outcomes. Typical examples of outcome-based predictions are the class of risk of a given execution or to the fulfilment of a predicate along the lifecycle of an execution trace~\citep{Maggi:CAiSE2014,Di-Francescomarino:2016aa,Leontjeva2015,Teinemaa2019};
    \item \textit{next activity predictions}, i.e., predictions related to the activities that are going to be executed. For example, it is possible to predict the sequence of activities of a process execution from the current time upon its completion~\citep{DBLP:conf/caise/TaxVRD17,DBLP:conf/bpm/Camargo19,Evermann2017}.
\end{enumerate}

Among the tools available that implement PPM techniques, the most used and known ones are: ProM~\citep{DBLP:conf/apn/DongenMVWA05}, Apromore~\citep{DBLP:journals/eswa/RosaRADMDG11}, and Nirdizati~\citep{DBLP:conf/bpm/RizziSFGKM19}. While the first two are well-known general-purpose tools covering different branches of the process mining field, the latter is specifically focused on PPM.

\subsection{Explainability Approaches}
\label{sec:explainability}
The lack of transparency of the black-box ML approaches, which do not foster trust and acceptance of ML algorithms, motivates the need of explainability approaches. In the literature, there are two main groups of techniques used to develop explainable systems, a.k.a.\ \emph{explainers}: post-hoc and ante-hoc techniques. Post-hoc techniques allow models to be trained as usual, with explainability only being incorporated at testing time. Ante-hoc techniques entail integrating explainability into a model since from the training phase. In this work, we mainly focus on post-hoc explainers since we want to use these instruments to improve the state-of-the-art approaches for PPM available in the literature without altering them but building new solutions on top of them.

An example of post-hoc explainer is Local Interpretable Model-Agnostic Explanations (LIME) \citep{DBLP:conf/kdd/Ribeiro0G16}, which explains the prediction of any classifier in an interpretable manner. LIME learns an interpretable model locally around the prediction and explains the predictive models providing individual explanations for individual predictions. 
The explanations are generated by approximating the underlying model with an interpretable one, learned using perturbations of the original instance.
In particular, each feature is assigned with an \emph{importance value} that represents the influence of that feature on a particular prediction. 
Another post-hoc explainer is SHapley Additive exPlanations (SHAP) \citep{DBLP:conf/nips/LundbergL17}. SHAP is a game-theoretic approach explaining the output of any ML model. It connects optimal credit allocation with local explanations using the classic Shapley Values~\citep{Shapley} from game theory and their related extensions. SHAP provides local explanations based on the outcome of other explainers thus representing the only possible consistent and locally accurate additive feature attribution method based on expectations.
Other post-hoc explainers~\citep{friedman2001greedy} show the marginal effect of some features using partial dependence plots. In~\citep{goldstein2015peeking}, the authors refine the partial dependence plots definition providing the visualisation of functional dependencies for individual observations through Individual Conditional Expectation (ICE) plots.
\citep{DBLP:conf/aaai/Ribeiro0G18} is also a model-agnostic system that explains complex systems using rules called anchors. The anchors, when available, are local conditions that can explain predictions. 

\subsection{Explainability in Predictive Process Monitoring}
\label{sec:explainability_ppm}
Recently, explainability approaches have also been applied and investigated in the field of PPM~\citep{harl2020explainable,DBLP:conf/bpm/SindhgattaMOB20,DBLP:journals/corr/abs-2008-07993,DBLP:conf/icpm/GalantiCLCN20,DBLP:conf/icsoc/Velmurugan0MS21,DBLP:conf/bpm/RizziFM20}. 

Some works focus on applying model-specific explainability approaches to provide explanations for predictions obtained through neural network predictive models, e.g., gated graph neural networks~\citep{harl2020explainable}, attention-based LSTM models~\citep{DBLP:conf/bpm/SindhgattaMOB20}, layer-wise relevance propagation to LSTM ~\citep{DBLP:journals/corr/abs-2008-07993}.

Other works focus on generic or model-agnostic post-hoc explanation approaches~\citep{DBLP:conf/icpm/GalantiCLCN20,DBLP:conf/icsoc/Velmurugan0MS21,DBLP:conf/bpm/RizziFM20}.
For example, in~\citep{DBLP:conf/bpm/RizziFM20}, explanations are used in order to identify the features leading to wrong predictions in order to improve the accuracy of the predictive model. In~\citep{DBLP:conf/icpm/GalantiCLCN20,GalantiICPM2021Demo}, shapley values~\citep{Shapley} are leveraged for providing users with explanations in the form of tables explaining the predictions related to a specific ongoing process execution. In addition, they use plots relying on heatmaps to specify for each feature and at each point in time the impact of the feature on a prediction. In Nirdizati~\citep{DBLP:conf/bpm/RizziSFGKM19}, different types of explanation plots are implemented to provide explanations in the context of binary classifications. In particular, explanation plots at \emph{event}, \emph{trace} and \emph{event log} levels are provided. For event-based explanations, LIME~\citep{DBLP:conf/kdd/Ribeiro0G16} and SHAP~\citep{DBLP:conf/nips/LundbergL17} explanation plots are adapted to the PPM scenario to measure the importance of each feature for a prediction provided at a specific trace prefix (see \figurename~\ref{fig:plotp1}). In order to provide trace-based explanations, a temporal stability~\citep{DBLP:conf/caise/Velmurugan0MS21,DBLP:journals/datamine/TeinemaaDLM18} plot is used. This plot allows users to visualize the importance of each feature at different trace prefixes (see \figurename~\ref{fig:plotp2}), thus also revealing how stable the importance of the features is for predictions returned at different prefix lengths. Finally, ICE~\citep{goldstein2015peeking} explanation plots have been implemented in Nirdizati to provide log-based explanations (see \figurename~\ref{fig:plotp3}). This type of plot reports, for a specific feature, information about the average value of the predicted label (between 0 and 1 with 0 meaning \emph{false} and 1 meaning \emph{true}) for the different feature values, as well as the number of traces in the event log containing each value.

    
\section{Related Work}
\label{sec:relatedwork}

Due to the rapid growth of the field of XAI and to the key role of users for these systems, several empirical studies with human subjects have been conducted to investigate and evaluate XAI systems. 


We can roughly classify the human-centered evaluation studies on explainability based on the specific AI approach for which explanations have been provided, e.g., neural networks, intelligent agents, random forest, generic ML, as well as on the form in which explanations are provided, e.g., textual, numeric or visual explanations~\citep{VILONE202189}. We mainly focus here on user evaluations dealing with explainability approaches developed for machine and deep learning techniques and providing a visual support to the users, which are the groups of works that are closer to our analysis.

We can classify evaluations for these XAI approaches based on the levels of tests used (test of \emph{satisfaction}, test of \emph{comprehension} or test of \emph{performance}), as well as the type of tasks the participants carried out (\emph{verification}, \emph{forced choice}, \emph{forward simulation}, \emph{counterfactual simulation}, \emph{system usage} or \emph{annotation})~\citep{Chromik2020ATF}. The evaluation can indeed be focused on evaluating the satisfaction of the users with the explanations or their subjective assessment of the understanding of the system; their comprehension of the system in terms of the mental model they have of the system; or the overall performance in terms of human-XAI system performance. In addition, in verification tasks, participants are asked about their satisfaction with the explanations; in forced choice tasks, the user is asked to choose among different explanations; in forward simulation tasks, participants are provided with data and explanations and asked to predict the system's output; in counterfactual simulation tasks, the participants should predict what input changes are required for getting an alternative output from the system; in system usage tasks, participants are asked to use the system for its original purpose, e.g., for decision making tasks; annotation tasks require participants to provide an explanation based on the input provided to the system and the produced output.


Some of the works in the literature focus on the evaluation of the only \emph{satisfaction} level. For instance,  in~\citep{DBLP:conf/chi/KrausePN16} an interactive interface (Prospector) is proposed to the users to understand how the features of a dataset affect the prediction of a model overall. A team of 5 data-scientists was asked to interact with this tool for 4 months to debug a set of models and, at the end of the experiment, the data scientists were interviewed on whether they feel that the provided support was beneficial for their work. In~\citep{DBLP:journals/jmui/WeitzSSHA21}, the effects of incorporating virtual agents for Explainable AI in a speech recognition system for keyword classification are investigated. The 60 participants were split into 4 groups: one group received only LIME explanation visualisations, while the other 3 groups were provided with additional information through different modalities from a virtual agent (text, voice and virtual presence). The results revealed a linear trend of the user's perceived trust: the visual presence of the agent combined with a voice output resulted in greater trust than the output of text or the voice output alone.
In~\citep{DBLP:journals/tvcg/SpinnerSSE20}, \emph{explAIner}, a visual analytic system for interactive XAI, is evaluated by 9 participants with different levels of expertise. The users were asked to use the system and their feedback was then collected. Results reveal that the system is intuitive and helpful in the daily analysis routine.


Other works focus not only on the evaluation of the satisfaction level but also of the users' \emph{comprehension}. 
For instance, in~\citep{DBLP:conf/nips/LundbergL17}, users were asked to carry out an \emph{annotation} task, i.e., they were asked to provide explanations on simple models. The human explanations were then compared with LIME, DeepLIFT and SHAP explanations and their consistency with human intuition evaluated. The results show that SHAP explanations are closer to human ones than the ones of the other approaches.
Also in~\citep{DBLP:conf/chi/TullioDCF07}, participants were asked to interact with a predictive system for two weeks at the end of which they were interviewed to collect their feedback and to check whether they gained some insights on the logic of the model to be explained by the explanatory system under analysis. Results show that participants were able to improve their mental model of the system by reducing initial misconceptions related  to the information used by the system, and better understanding how it works, also without technical knowledge.

Besides satisfaction and comprehension, a last group of works focuses instead on evaluating the \emph{performance} of the user (and the system). 
For instance, in~\citep{DBLP:journals/dss/HuysmansDMVB11}, the authors evaluate the levels of comprehensibility of decision table, decision tree and rule-based predictive models with a user study through \emph{forward-simulation} tasks. The user study involved 51 non-expert users who were asked to carry on classification tasks
and accuracy, response time and confidence for each of the three types of predictive models were then evaluated. The analysis of the results reveals that decision tables outperform significantly all the other methods.
Also in~\citep{DBLP:conf/iui/WangY21}, users were asked to carry on \emph{forward-simulation} tasks. The work compares and evaluates four common model-agnostic explainable AI methods on these tasks. The results show that the effect of AI explanations depend on the level of domain expertise of the users. 
Similarly, the empirical evaluation carried out in~\cite{Kulesza2015} focuses on comparing explanatory debugging and traditional black-box learning systems. 77 participants were asked to make predictions as close as possible to the ones provided by the system. The results show that participants using the explanatory debugging were able to understand the learning system 50\% better than the black-box system and to correct its mistakes.

Together with \emph{forward-simulation} tasks,  \emph{annotation} tasks were also proposed to 64 participants in~\citep{DBLP:conf/iui/AlqaraawiSWCB20}. The aim of the work was evaluating the performance of saliency maps generated by layerwise relevance propagation (LRP)~\citep{bach-plos15}. The results show that although saliency maps help users understand the features the system is sensitive to, they provide limited support for the classification of new instances.
Similarly, in~\citep{Cheng2019}, 199 participants used four different explanation interfaces to understand a decision-making algorithm. Participants were asked to perform some \emph{annotation} and \emph{counterfactual} tasks. The results show that interactive explanations improve the understanding of the algorithm, although it requires more time.

In~\cite{DBLP:conf/acl/HaseB20}, \emph{counterfactual simulation} tasks were given to participants, besides \emph{forward-simulation tasks}. In this work, the authors propose a user study to evaluate the \emph{simulatability} of different explanation approaches, where an approach is simulatable if a user is able to predict its behavior on new instances.

Finally, in some works, participants are asked to carry on \emph{system usage} and, more specifically, decision-making tasks. 
For instance, in~\citep{DBLP:conf/kdd/Ribeiro0G16}, an evaluation with human subjects is carried out with the two explanation techniques proposed in the work, i.e.,  LIME and SP-LIME. Experts and non-expert participants were required to carry on four decision making tasks, i.e., (i) deciding whether to trust a prediction; (ii) choosing among different classification models; (iii) carrying on feature engineering; (iv) identifying why a classifier should not be trusted. Results show that explanations are actually useful to support users in these tasks.
Similarly, in~\citep{DBLP:conf/atal/MalhiKF20}, 65 participants were asked to carry out a decision-making task in order to evaluate the human understanding of the behavior of explainable agents. Three user groups were considered: a group was provided with an agent without explanations and the other two with explanations generated by LIME and SHAP, respectively. The results show that notable (though not statistically significant) differences exist between the groups without and with explanations in terms of bias in human decision making. 
Decision-making tasks are also used in~\citep{krause2018user} to evaluate the effect of using aggregated rather than single data point explanations. The results show that using aggregated explanations has a positive impact on the detection of biases in data.

To the best of our knowledge, however, so far, no user evaluation has been conducted on how users understand and use explanation plots in the field of PPM. To this aim, in this work, we evaluate explanation plots at \emph{satisfaction}, \emph{comprehension} and \emph{performance} level.
\section{Methodology}
\label{sec:method}
To study how individuals make sense of explanation plots (\hr{RQ1}), how they use them for decision making (\hr{RQ2}) and identify means to improve them (\hr{RQ3}), we conducted a qualitative observational study. This approach is suitable because it allows us to draw insights into ``how'' individuals interact with prediction explanations in the context of PPM~\citep{lazar2017research}.

Here, we first provide an overview of our study setting (Section~\ref{sec:method:setting}) before discussing our data collection strategy (Section~\ref{sec:method:data}) and analysis methodology (Section~\ref{sec:method:procedure}).

\subsection{Setting}
\label{sec:method:setting}
For our study, we selected eight individuals with different backgrounds as participants. We focused on individuals that had expertise in Business Process Management (BPM) since such individuals would commonly make decisions related to how processes are conducted. We selected four individuals that had expertise in BPM and ML, and four individuals that had expertise in BPM only. We made this differentiation since it can be expected that knowledge related to ML would affect whether and how individuals understand explanation plots and how they use them for decision making. Within these two groups of four participants, we selected individuals from different domains to foster the applicability of our findings beyond the confines of a specific use case.



For setting our study, we decided to create two separate scenarios: one from the medical domain (Section~\ref{sec:method:setting:domaina}) and one from the financial domain (Section~\ref{sec:method:setting:domainb}). We selected these domains because they are common domains where PPM techniques are applied~\citep{vandongen_2011,vandongen_2012,vandongen_2017}.
We opted for developing scenarios based on real-life settings rather than conducting the study \emph{in situ} for three main reasons. First, PPM methods are not widely used in practice yet, which makes it difficult to find suitable test beds. Second, studying such methods in a real setting would almost certainly lead to domain knowledge of individuals to impact our findings. Third, using a real setting instead of an artificial one would make it difficult to compare findings of different study subjects. In the following, we describe the two scenarios we used (Sections \ref{sec:method:setting:domaina} and \ref{sec:method:setting:domainb}) before elaborating on the plots we used (Section~\ref{sec:method:setting:plots}) and the tasks that we asked participants to complete within the two scenarios (Section~\ref{sec:method:setting:tasks}).

\subsubsection{Domain A (medical)}
\label{sec:method:setting:domaina}
We consider a process pertaining to the treatment of patients with fractures, for which we want to predict patients who will recover soon or late. Every process execution starts with examining the patient. If an X-ray is performed, then the X-ray risk must be assessed before it. Treatments reposition, cast application and surgery require that an X-ray is performed before. If a surgery is performed, then a rehabilitation must be prescribed eventually after it. Finally, after every cast application a cast removal must be performed. Each process execution refers to a patient whose age is also known. Moreover, for each patient the type of treatment prescribed is also known and whether or not the patient should carry on a rehabilitation after the treatment. We predict whether the patient will recover quickly (1) or not (0).

\subsubsection{Domain B (financial)}
\label{sec:method:setting:domainb}
We consider a process of a bank to handle the closing of a bank account. A request is first created, by either the owner of the account or a 3rd-party, such as an attorney. Then, the request is evaluated. As part of the evaluation, a risk assessment may also occur, which involves checking for abnormal transactions in the account history, or whether the account has been involved in illicit or suspicious activities. Risk assessment is optional and may also be executed later in the process. Then the outstanding balance of the account is determined, including pending payments. Before the closing can be finalized, an investigation of the account owner’s heirs has to be executed, in order to understand to whom the outstanding balance should be transferred. The outcome of the process can be either of the following: the request will be executed (1) or the request will be sent to credit collection (0) for a more thorough assessment. The latter may happen, for instance, when there are irregularities in a request, pending payments, or the heirs are unreachable. Since this procedure takes a long time and involves a high amount of resources, the bank would like to minimise the number of requests sent for credit collection.

\subsubsection{Plots}
\label{sec:method:setting:plots}
We selected three plots for our study in order to investigate three levels of explanations for predictions: (i) at \emph{event}, (ii) \emph{trace} and (iii) \emph{event log} level.

\begin{figure}[!ht]
    \centering
    \includegraphics[width=0.9\linewidth]{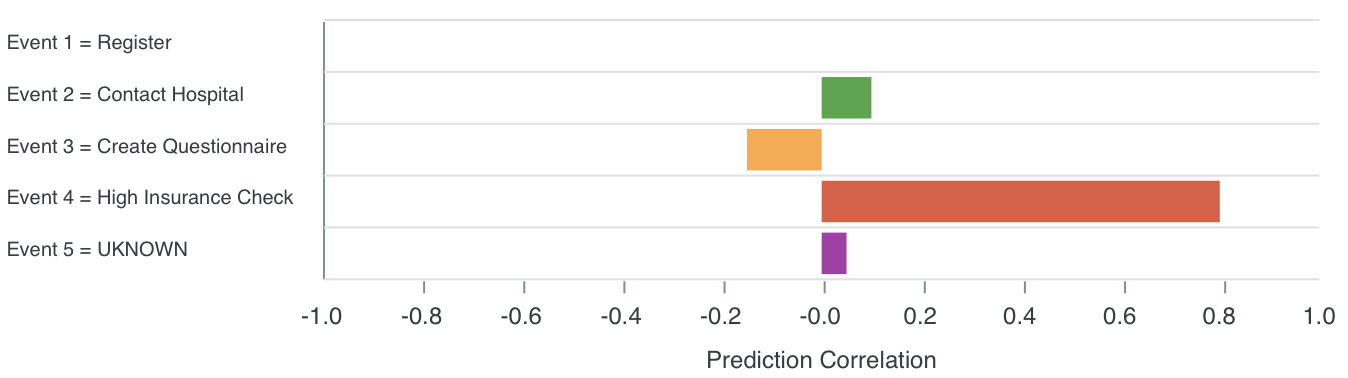}
    \caption{SHAP values related to the correlation between the pair (feature,value) and the returned prediction (\emph{Plot P1})}
    \label{fig:plotp1}
\end{figure}

The first plot (\emph{Plot P1}) - the plot at \emph{event} level - is an adaptation of the SHAP explanation plots to the PPM field. \figurename~\ref{fig:plotp1} shows an example of this type of plot. For all features, which in the case of an encoding based on the event position in the trace (index-based encoding~\citep{Leontjeva2015}) correspond to the activities executed at different positions of the trace, the plot shows the impact on the prediction returned by the predictive model, i.e., the correlations in terms of SHAP values of each feature (and associated value) with the prediction. For instance, in \emph{Plot P1}, reported in \figurename~\ref{fig:plotp1}, the most impacting feature-value pair is the pair composed of the event at position four (\feat{event\_4}) and activity \act{High Insurance Check}, i.e., activity \act{High Insurance Check} occurring at position four has a high correlation with the returned prediction (expressed as a SHAP value of $0.8$). Such a correlation means that this feature-value pair strongly affects the returned prediction.

\begin{figure}[!ht]
    \centering
    \includegraphics[width=0.8\linewidth]{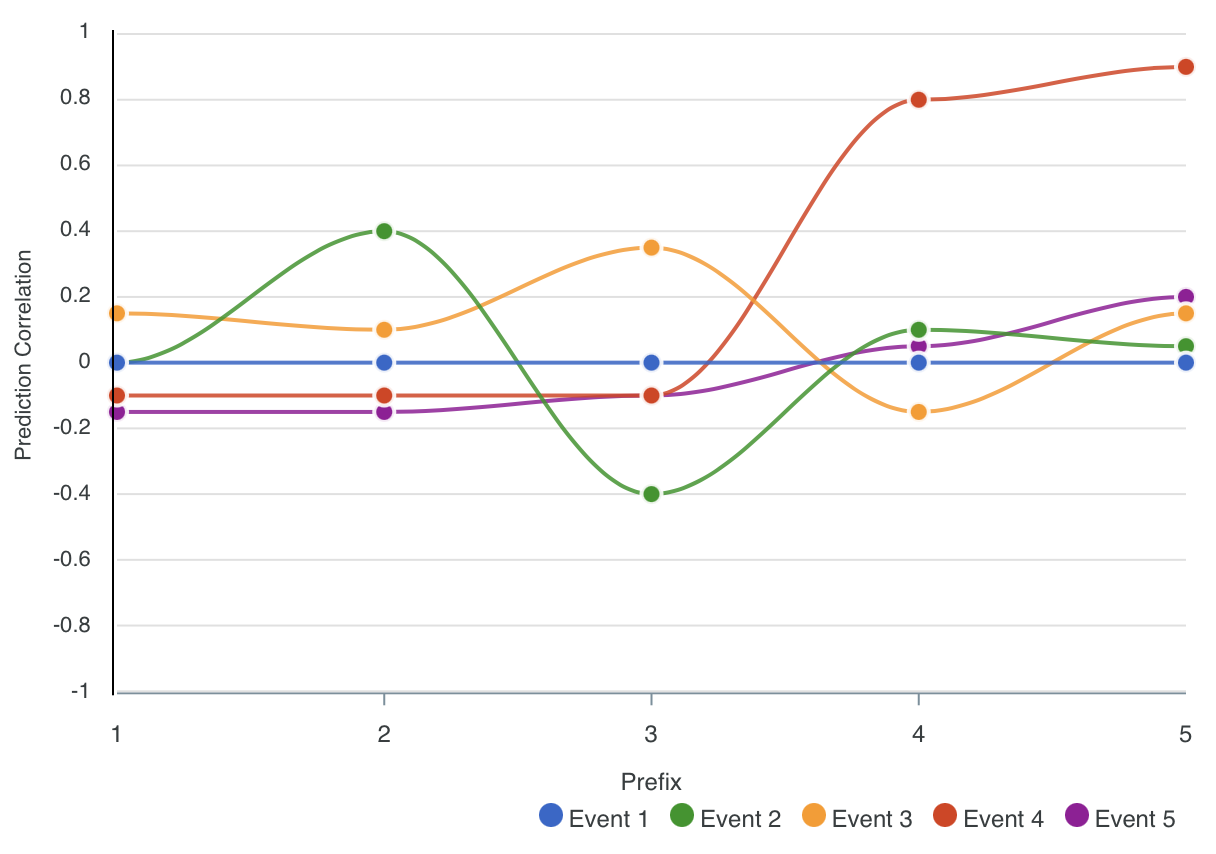}
    \caption{Temporal stability of the SHAP values related to the correlation between each feature and the returned prediction (\emph{Plot P2})}
    \label{fig:plotp2}
\end{figure}

The second plot (\emph{Plot P2}) - the plot at \emph{trace} level - shows how the correlation of each feature with the returned prediction (expressed in terms of SHAP values) changes at different positions in the ongoing trace. The plot shows hence how \textit{stable} the importance of a feature for a prediction is as the chosen prefix length at which the prediction is carried out increases. The plot is an adaptation of the \emph{temporal stability} plot~\citep{DBLP:journals/datamine/TeinemaaDLM18} used in PPM to check the stability of a prediction at increasing prefix lengths to the case of the prediction explanations. Differently from the original temporal stability plots, in which the prediction values are plotted for different prefix lengths, in these plots, the SHAP values of the features (and corresponding values) are plotted for different prefix lengths. For instance, in \emph{Plot P2}, reported in \figurename~\ref{fig:plotp2}, feature \feat{event\_2} corresponding to the event at the second position of the trace (green line) is rather unstable, as it positively or negatively correlates with the returned prediction based on the specific point of the trace in which the prediction is made. Instead, feature \feat{event\_4} corresponding to the event at position four of the trace (red line) positively correlates with the prediction starting from the prediction carried out at prefix length four in a stable way. This means that, starting from prefix of length four, the returned prediction highly depends on the event occurred at position four, which is actually known only from prefix length four on. Note that, in the plot in \figurename~\ref{fig:plotp2}, the specific values of the features are not reported, as they may change for different prefix lengths, as in the case of \feat{event\_4} at prefix four whose value for prefixes shorter than four is set to null. The values of the features are however visible by hovering the mouse on the dynamic version of the plot (see \figurename~\ref{fig:appendix:hoverP2}). The plot allows the user to get an idea of the stability of the importance of a certain feature as the trace evolves, thus increasing the confidence of the user in making decisions based on explanations using that feature~\citep{goldstein2015peeking}.

\begin{figure}[!ht]
    \centering
    \includegraphics[width=.9\linewidth]{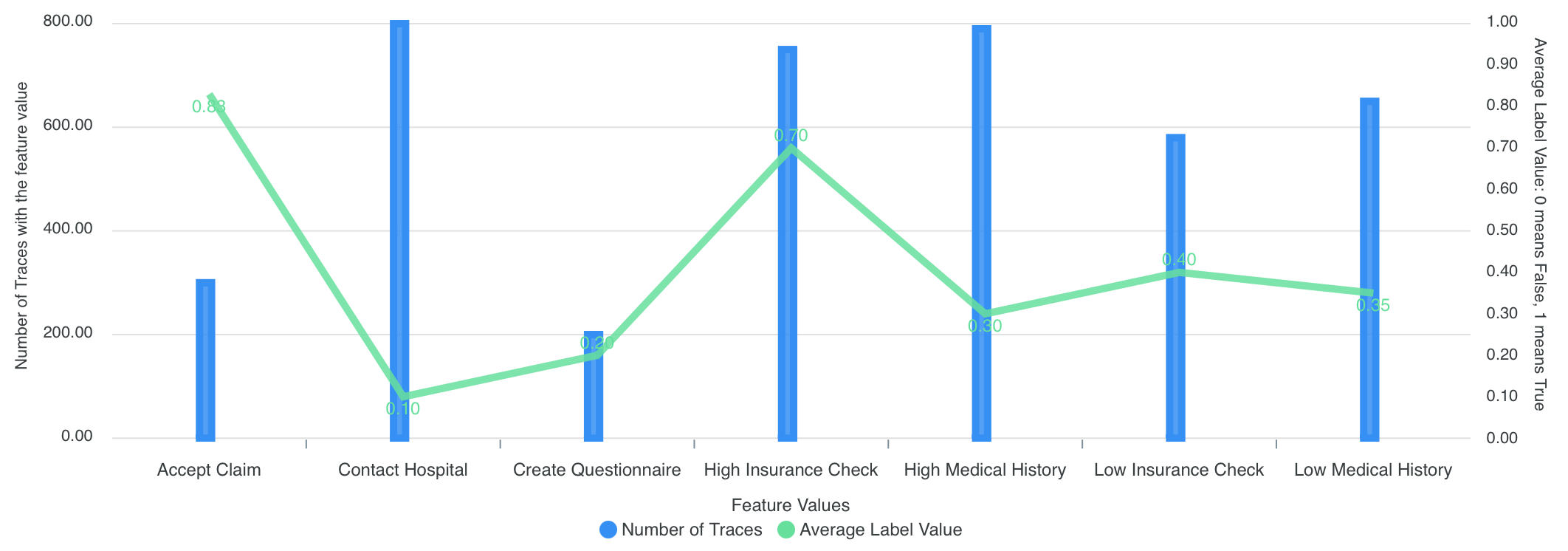}
    \caption{Average target value through different values of a feature on the whole event log (\emph{Plot P3})}
    \label{fig:plotp3}
\end{figure}

The third plot (\emph{Plot P3}) - the plot at the \emph{log} level - shows, for a specific feature, the average value of the predicted label (between 0 and 1 with 0 meaning \emph{false} and 1 meaning \emph{true}) for the different feature values, as well as the number of traces in the event log containing each value. The plot, which mainly collects some statistics on the training data, is based on the ICE plots~\citep{goldstein2015peeking}. In a binary setting, in which two possible values are associated to the predicted label, the average value of the label is the average percentage of traces for which the label value is \lab{true}. For example, \emph{Plot P3}, reported in \figurename~\ref{fig:plotp3}, shows the average value of the predicted label for the different values of feature \act{event\_4}, as well as the frequency of the values of that feature on the whole event log. For instance, value \act{High Insurance Check} for feature \feat{event\_4} has an average label value of $0.7$, i.e., in $70\%$ of the traces of the training set having as value for feature \feat{event\_4} activity \act{High Insurance Check}, the associated label is \lab{true}, while, for the remaining $30\%$, the label is \lab{false}. Moreover, the plot also shows that such a value occurs in around $750$ traces in the training set. Therefore, the plot allows the user to get an overall idea of the distribution of the labels for a specific feature value, as well as of the frequency of that specific value in the whole event log.

\subsubsection{Tasks}
\label{sec:method:setting:tasks}
For each of the aforementioned scenarios (Sections~\ref{sec:method:setting:domaina} and \ref{sec:method:setting:domaina}), we developed a set of tasks that required participants to use the three types of plots described in Section~\ref{sec:method:setting:plots} for decision making.

\begin{figure*}[!h]
    \centering
    \begin{subfigure}[b]{0.49\textwidth}
        \centering
        \includegraphics[width=\linewidth]{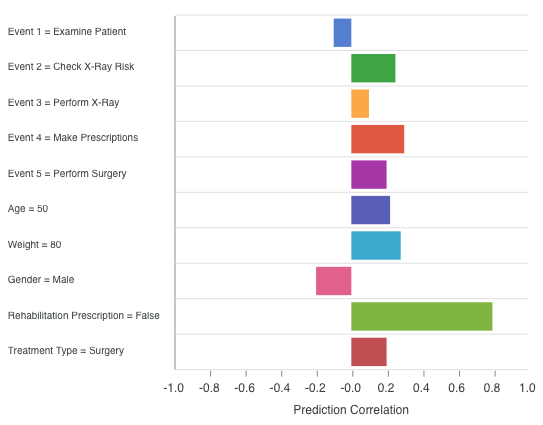}
        \caption{Predicted outcome for late recovery.}
        \label{fig:method:setting:tasks:plotA1a}
    \end{subfigure}
    \hfill
    \begin{subfigure}[b]{0.49\textwidth}  
        \centering
        \includegraphics[width=\linewidth]{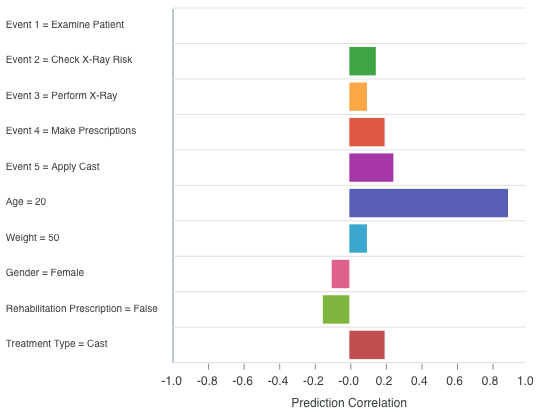}
        \caption{Predicted outcome for quick recovery.}
        \label{fig:method:setting:tasks:plotA1b}
    \end{subfigure}
    \caption{SHAP values related to the correlation between features, their value and the predicted outcome (\emph{Plot P1}).} 
\end{figure*}

In the following, we will describe the tasks we developed for domain A (Section~\ref{sec:method:setting:domaina}). The full list of tasks including the domain description, plots and the questions the investigator asked can be found in Appendix~\ref{sec:appendix:domains}:

\textbf{Task Description A1.a:} Consider an incomplete trace of a patient who has carried out one of treatments reposition, cast application, or surgery. For this patient, the prediction is that she will not recover soon from the fracture (0). The explanation of the prediction is reported in Figure~\ref{fig:method:setting:tasks:plotA1a}.

\textbf{Task Description A1.b:} Consider now the incomplete trace of another patient who has carried out one of treatments reposition, cast application, or surgery. For this patient the prediction is that she will recover soon from the fracture (1). The explanation of the prediction is reported in Figure~\ref{fig:method:setting:tasks:plotA1b}.

\begin{figure}[!h]
    \centering
    \includegraphics[width=.9\linewidth]{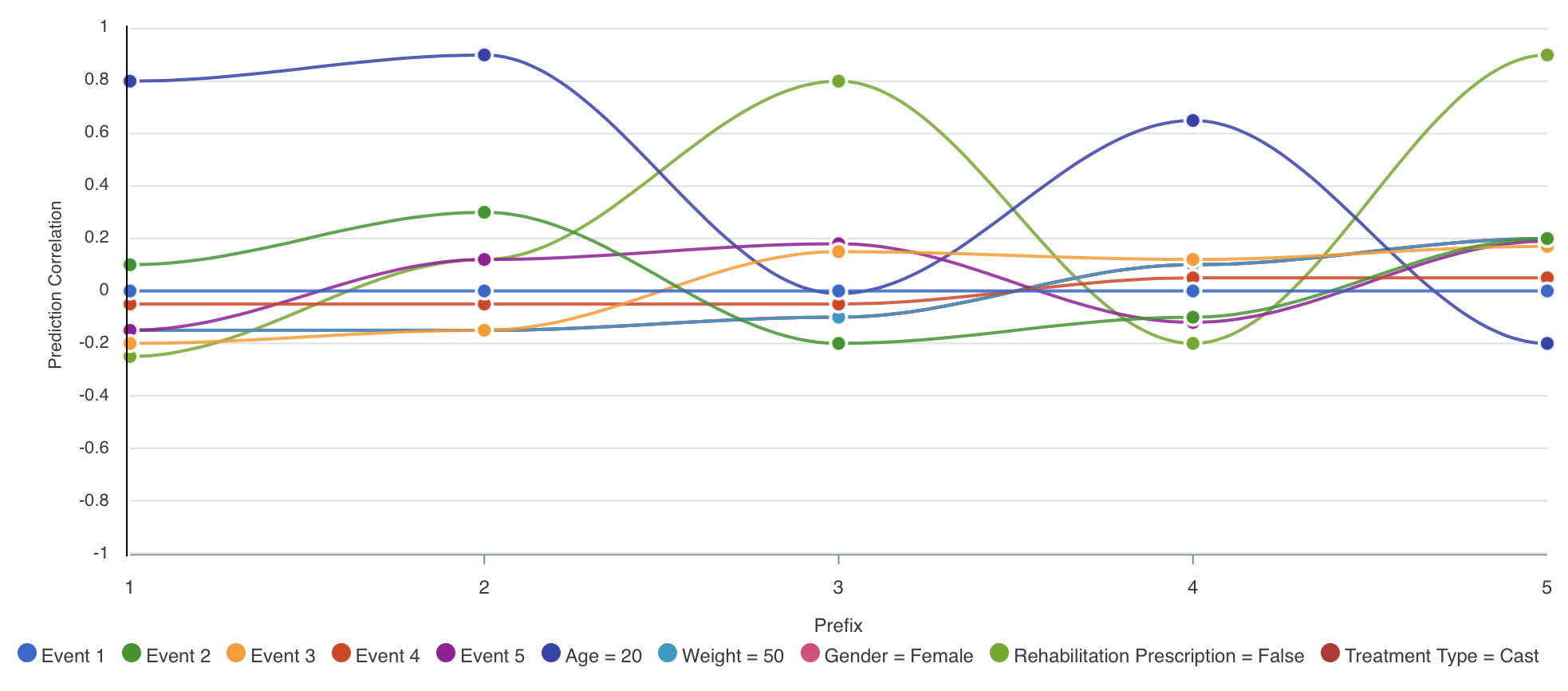}
    \caption{Temporal stability of the SHAP values related to the correlation between each feature and the predicted outcome for a quick recovery (\emph{Plot P2}).}
    \label{fig:method:setting:tasks:plotA2}
\end{figure}

\textbf{Task Description A2:} Consider an incomplete trace of a patient who has carried out one of treatments reposition, cast application, or surgery. For this patient, the prediction is that she will recover soon from the fracture (1). The explanation of the predictions from the beginning of the trace up to the current point is reported in Figure~\ref{fig:method:setting:tasks:plotA2}.

\begin{figure}[!h]
    \centering
    \includegraphics[width=.7\linewidth]{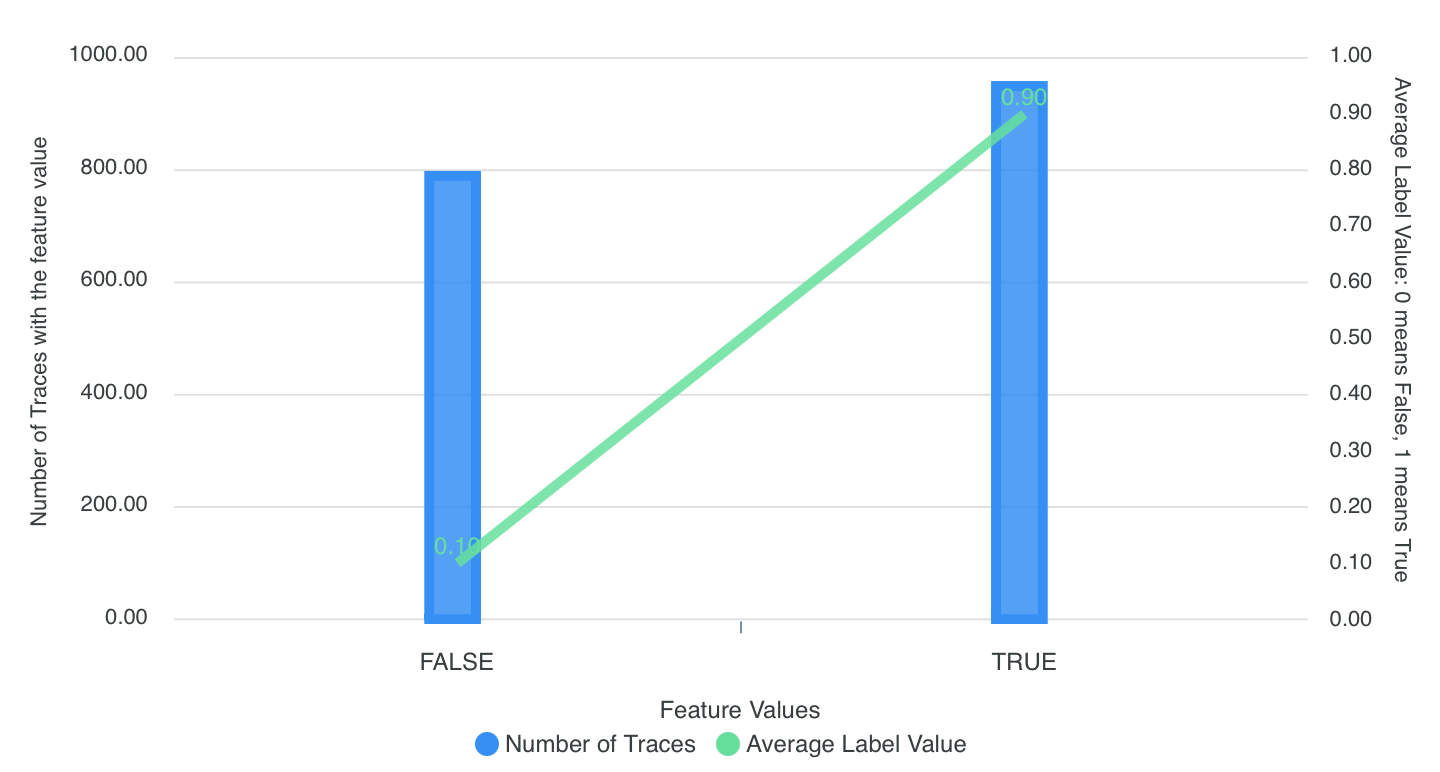}
    \caption{Average recovery time from a fracture through different values of a feature on the whole event log (\emph{Plot P3}).}
    \label{fig:method:setting:tasks:plotA3}
\end{figure}

\textbf{Task Description A3}: Consider a set of process executions related to patients who have carried out one of treatments reposition, cast application, or surgery. For some of these patients, the prediction is that they will recover soon from the fracture (1), for others the prediction is that it will take time for them to recover (0). The explanation of the predictions for different patients in the training set is reported in Figure~\ref{fig:method:setting:tasks:plotA3}.

In addition, we also developed a comprehension task that we used at the beginning of each individual observation. The aim of this task was for participants to be able to familiarize themselves with the format of the study as well as with the plots used.

\subsection{Data collection}
\label{sec:method:data}
We conducted semi-structured observational interviews with individuals within our study population which lasted between 68 and 109 minutes each. To guide the participants through the study procedure, we created a web interface that contained the description of the domains as well as the related plots and tasks\footnote{The interface can be accessed at \url{https://user-evaluation-mock.web.app/}}.

The first part of the web interface contained explanation plots and tasks for the comprehension task. The second and third parts contained explanation plots and tasks for domains A and B. We presented the two scenarios in different orders within the two groups of subjects to be able to mitigate the effect of individuals learning how to interpret the plots (c.f.\ Table~\ref{tab:experts:split}).

\begin{table}[!h]
\centering
\begin{tabular}{ccc}
Expertise & Domain Order & Code   \\ \hline
\multirow{4}{*}{\bpm}      & \multirow{2}{*}{A then B} &  BA1\\
      &  &  BA2\\\cline{2-3}
      & \multirow{2}{*}{B then A} &  BB1\\
      &  &  BB2\\
      \hline
\multirow{4}{*}{\bpmml} & \multirow{2}{*}{A then B} &  MA1\\
 &  &  MA2\\\cline{2-3}
 & \multirow{2}{*}{B then A} &  MB1\\
 &  &  MB2 \\ \hline
\end{tabular}
\caption{Participants, their expertise and provided order of domains}
\label{tab:experts:split}
\end{table}

The interviews were conducted via Zoom by a team consisting of a \emph{facilitator} and an \emph{observer}, with the facilitator guiding the participant and the observer serving in a supporting role. Participants were encouraged to think aloud, to ask questions and to point out interesting aspects related to the different plots during the interview. All interviews were video-recorded. 

Each interview started with the facilitator introducing the study procedure and the application to the participants who were then asked to open the link to the comprehension task and share their screen. The facilitator then asked the participants a set of predefined questions regarding their understanding (e.g., ``\emph{What do the different plots show?}'') and interpretation (e.g., ``\emph{Which are the feature(s) influencing most the prediction related to the visualized trace?}'') of the different plots (\hr{RQ1}).

After the comprehension task the participants were asked to open the web interface containing explanation plots and tasks for domains A and B. The facilitator then introduced domains and tasks and asked the participants to explain the plots (e.g., ``\emph{How do you interpret the plot?}'', \hr{RQ1}) and to decide what to do next based on their interpretation of the plot (e.g., ``\emph{Based on the information presented in the plot, what course of action would you suggest?}'', \hr{RQ2}). In addition, the facilitator also asked specific questions related to each plot (e.g., ``\emph{Would you recommend carrying on the rehabilitation? Why?}'', for task A1.a) to further examine the decision making process leading the participant to suggest a certain course of action (\hr{RQ2}).

After finishing the tasks, the facilitator conducted a short post-interview. The questions focused on the participant's understanding of the different plots (e.g., ``\emph{Which visualization(s) was/were the hardest to interpret? Why do you think you were struggling with these visualizations in particular?}'', \hr{RQ1}) and the perceived usefulness for decision making (e.g., ``\emph{Which of these visualizations helped you to make more informed decisions?}'', \hr{RQ2}). In addition, the facilitator also asked each participant suggestions on how to improve the plots to better support their ability of making informed decisions based on them (e.g., ``\emph{Is there any additional information that would have helped you to understand the different visualizations and make more informed decisions during the study?}'', \hr{RQ3})

Finally, the participants were asked to answer a questionnaire after the interview. The questionnaire focused on the perceived ease of use (\hr{RQ1}) and perceived usefulness of each plot (\hr{RQ2}). For this, we adapted common scales included in the technology acceptance model~\citep{davis1989perceived}. In addition, we asked participants to rate their experience related to BPM, ML, PPM and Explainable AI on a scale from 1 to 10.

\subsection{Analysis procedure}
\label{sec:method:procedure}
Our qualitative analysis mainly focused on the video recordings, observations and follow-up interviews. The collected questionnaire data served as an additional qualitative data point.

For our analysis, we combined deductive and inductive coding. We started with a set of predefined codes related to our research questions. We focused on whether and how participants made sense of the different plots (\hr{RQ1}) using codes such as \emph{correct interpretation}, \emph{wrong interpretation} and \emph{required additional information} and on how participants used the plots for decision making using codes such as \emph{correct decision}, \emph{wrong decision} and \emph{reasoning}. We also included codes focusing on the improvement of the different plots such as \emph{labeling} and \emph{interaction with visualizations}.

A group of three researchers all of which are co-authors of this paper collaboratively applied these codes to all interviews. The coding was done by plot meaning that first all responses related to plot 1 for all interviewees were coded before moving on the plot 2. Afterwards the same researchers conducted a second round of inductive coding adding aspects that emerged during the first round analysis including \emph{confidence}, \emph{trust in visualization} and \emph{what-if-analysis}. These codes were then again collaboratively applied by the three researchers. The last step was that the researchers clustered the findings into themes related to our three research questions.

\section{Findings}
\label{sec:findings}
In this section, we discuss our findings related to how participants made sense of explanation plots (Section~\ref{sec:findings:rq1}, \hr{RQ1}) and how they utilized them to make decisions (Section~\ref{sec:findings:rq2}, \hr{RQ2}). We also outline suggestions on how plots can be improved (Section~\ref{sec:findings:rq3}, \hr{RQ3}).

\subsection{RQ1: How do users make sense of explanation plots?}
\label{sec:findings:rq1}





\begin{figure}[h]
    \centering
    \includegraphics[width=.7\linewidth]{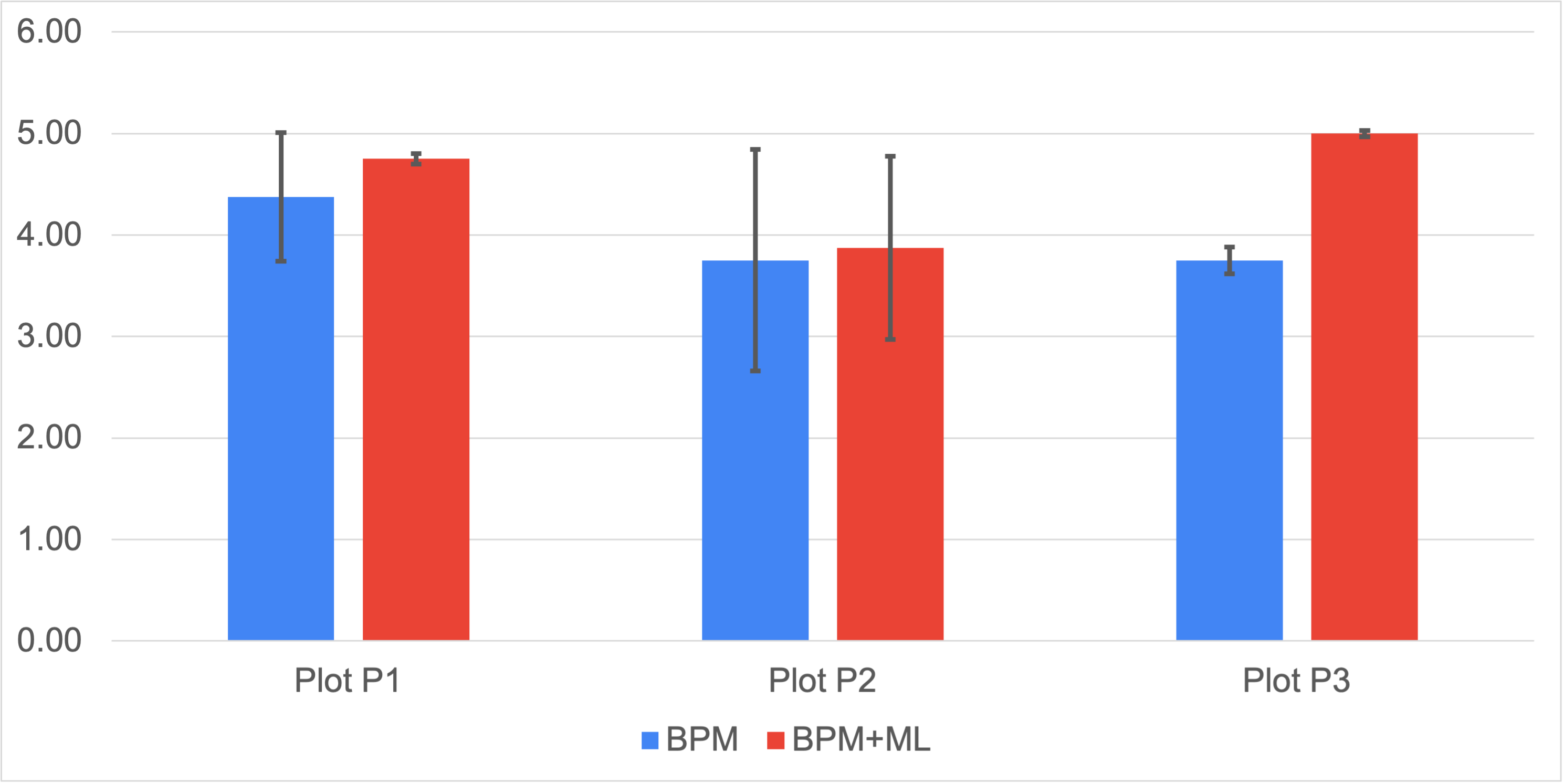}
    \caption{Perception of \bpm and \bpmml experts related to the ease of use of the different explanation plots (1=very low to 5=very high scale). We report in the plot also median and variance.}
    \label{fig:survey_peou}
\end{figure}

Our interviewees generally found the different plots to be reasonably easy to use. This is evident by survey responses we received (Figure~\ref{fig:survey_peou}) and also by some statements of different participants during the interviews (\emph{“I found [these] quite easy to interpret”}, \anti, \emph{“It's easy to get”}, \claudio, \emph{“the interpretation is clear”}, \francesco). 
Out of the three plots we used, the participants perceived P1 to be the easiest and P2 to be the hardest to use. The \bpmml experts we interviewed generally perceived the plots to be easier to use than the \bpm experts. This difference is minimal for plots P1 and P2 while albeit slightly larger for P3.

The fact that plots were generally easy to use, however, does not mean that the experts did not have to climb a learning curve to fully master the use of the plots. The main themes regarding the difficulties in understanding and using the plots emerged during the interviews with both types of experts are discussed next. 

\subsubsection{Complicated first impressions}

Most users struggled at the beginning with understanding specific aspects of each plot. This could have been expected, considering that many experts, in particular the \bpm ones, were seeing such plots for the first time.

In P1, the main misunderstanding from experts concerned the meaning of the Y-axis. The Y-axis in this plot simply lists the features that are considered by the model at a certain prefix and the order in which these are shown does not matter. During the interviews, however, some experts associated a meaning with the ordering of the features on the Y-axis. \anti and \claudio associated this with a notion of feature importance, i.e., the higher features on the Y-axis, were considered somehow more important
in the predictive model (\emph{“So this is from top to bottom, I'm assuming based on, okay, some kind of value for each event of the event log”}, \anti). \han interpreted this as a chronological order, i.e., the higher features on the Y-axis, were assuming the corresponding values earlier (\emph{"I think that after event one, there is nothing yet.... And then after event [...] it seems that there is a relatively low, but definitely some positive correlation"}, \han). Such misinterpretations may also have been due to the fact that the comprehension task only used features defined by event labels.

Regarding P2, two \bpmml experts (\han and \francesco) struggled to understand the fact that the  data shown are part of a time-series and that, therefore, they have to be considered as a whole. That is, the correlation values are valid at a given prefix length, and they may vary if the same plot is regenerated for a different prefix length 


Finally, the \bpm experts considered P3 too complex. For \adela and \claudio, the plot was hard to understand at the first glance because of its different graphical appearance and different semantics compared to plots P1 and P2.  The double Y-axis, representing the average label value and the number of traces with that feature value, also was deemed as confusing by the same experts (``\emph{\textit{Expert}: So, trying to find out what for instance, the green value means it is average labor value or contact hospitals which means all traces that had an event contact hospitals. \textit{Interviewer}: that have contact hospital at event 4 is the average value of the label. \textit{Expert}: Okay, and now I get it.}'', \adela; ``\emph{\textit{Expert}: With the average label value, average label value. [...] What is an average label value? \textit{Interviewer}: Labels can be one or zero. \textit{Expert}: Yeah, the outcome. Right.}'' \claudio).

Except these issues, after struggling initially, the experts managed to correctly interpret the information in the plot without explicit help from the facilitator.




\subsubsection{Additional explanations required}

In many cases, the experts explicitly required additional explanations from the facilitator to understand specific aspects of each plot.

Some experts in the \bpm group required an explicit explanation of the meaning of ``correlation'' in plot P1 (\emph{"The prediction correlation to be honest, it doesn't tell me anything. So, there is a set of events and correlation with respect to what?"} \pierluigi).  Here, the lack of experience with explainable AI techniques of the \bpm experts plays a key role in their understanding of the targets of the correlation mentioned on the X-axis of P1.

After having seen P1, all experts were able to understand P2. However, some experts struggled to understand the meaning of what a ``stable feature'' in P2 is and required the explicit guidance of the facilitator. Intuitively, some experts associated feature stability with a \emph{flat} line in the plot. However, they also realized immediately that a flat line at 0 correlation identifies an irrelevant feature that has no correlation with the trace outcome at any prefix length (\emph{"The stability suggest me [one feature] because it's always zero. But it also means that doesn't affect at all the the prediction. It is stable but it's useless"} \pierluigi). After this initial hesitation, most experts (\pierluigi, \anti,\han, \francesco, and \manuel) correctly considered as stable the feature(s) for which the line in the plot was flat after having taken a positive value.   
Some of the experts, however, considered other aspects when determining stable features in P2, such as the trend of the correlation value (\anti, \han), e.g., \emph{"The correlation moving to negative value could mean the counterexample of the possible alternative feature to outcome"} (\anti) , and the relative change of correlation value (\francesco) to identify the stable features. 
Finally, one expert (\adela) incorrectly identified as stable only the feature with highest correlation value at the last prefix shown in P2.  



\subsubsection{Effect of data imbalance in P3}

During the interviews, the experts generally understood correctly the information provided in P3. For instance, \adela, who required explicit explanations on both P1 and P2, could easily  understand the content of P3, even though the information conveyed by this plot is far more than the one provided in P1 and P2 (``\emph{For me, it was clear that the correlation between the values of a feature that we are evaluating [...] There is like less noise, so to say, like less information that I'm not using to make my assumptions.}'', \adela).
However, two experts (\pierluigi and \francesco) assumed that P3 could point to a predictive model that is unreliable or biased when the labels in the event log are not balanced (``\emph{It could be more reliable whenever I have a set of traces that has an equal number of true and false.}'', \pierluigi; ``\emph{If you have a very skewed data set where you have too high number of traces having the same label and where the other label in a binary setting are fewer [...] then you can have some bias in the prediction.}'', \francesco). While there is no technical argument to support this claim, we argue that these experts over-analyzed the information shown in the plot.

\subsection{RQ2: How can explanation plots support users in decision making?}
\label{sec:findings:rq2}
\begin{figure}[h]
    \centering
    \includegraphics[width=.7\linewidth]{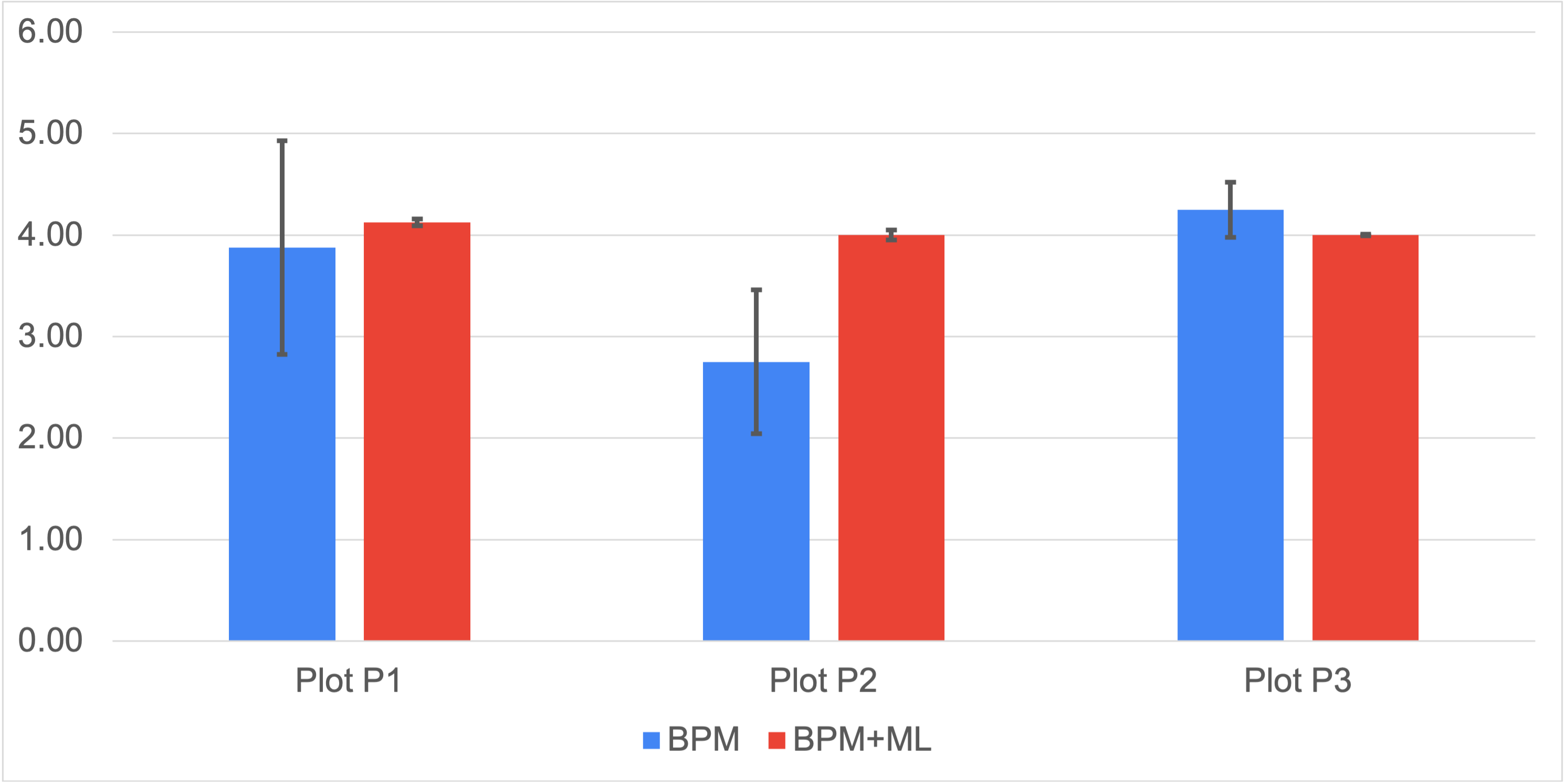}
    \caption{Perception of \bpm and \bpmml experts related to the usefulness of the different explanation plots (1=very low to 5=very high scale). We report in the plot also median and variance.}
    \label{fig:survey_pu}
\end{figure}

Similar to the findings reported in the previous section (Section~\ref{sec:findings:rq1}) our interviewees generally found the different plots to be reasonably useful for solving the decision making tasks we proposed as evident by the survey responses we received (Figure~\ref{fig:survey_pu}) and by some participant statements
(``\emph{I clearly understood what I should recommend}'', \adela, ``\emph{all of [the plots] serve a purpose}'', \anti, ``\emph{So it's very intuitive the way to represent this part here,}'', \claudio). 
The general perception about the usefulness of the plots is lower than the perception about their ease of use though. The survey also showed that the \bpmml experts who participated in our study found all three plots equally useful. The \bpm experts, however, perceived P2 to be considerably less useful than the other plots.

To answer RQ2, we have identified several common themes regarding how the experts have used the plots in the decision making tasks. The themes that have emerged are, in most cases, common to both \bpm and \bpmml experts. The themes specific to one particular expert group are discussed at the end of the section.  



\subsubsection{Correct decision, but wrong reasoning}

In some cases, experts ended up making the correct decision in a task, but interpreting the information in the plots in the wrong way.
A typical example of this situation is the one of experts (\adela, \anti, \han, and \francesco) using only the information associated with the last prefix shown in plot P2 in Task A2. It is only by coincidence that this led the experts to make the correct decision. The reason behind this behavior could be that experts anchored their reasoning on the principle that the more information is given to a predictive model, the more accurate it is likely to be (
``\emph{The latest you make the prediction, the more accurate it is because then you have more information}'', \adela). Hence, they considered reasonable to trust only the information associated with the longer prefix in plot P2.

Another example concerns experts suggesting to prescribe the rehabilitation in task A1 because it looked like the only option available, even though other options were available, like proceeding without taking any action (``\emph{I suggest doing [..] for the reason that it's the only thing that the [doctor] can change, or affect.}'', \anti). 

\subsubsection{Varying level of confidence in the decision made}
\label{sec:findings:confidence}

For different reasons, the experts were not always confident in the decisions they made using the plots. This was particularly the case of tasks in which plot P1 was used (Tasks A1 and B1).
Right at the end of task A1, when asked for general comments, \niek was concerned by the possibility of making a wrong choice  (``\emph{What is the real business risk and setting? What kind of decisions? [...] how do I need to act on the information I see?}'', \niek)  and, in the subsequent tasks, often refused to make decisions based on the information shown by the plots. 
%

\han, in both task A1 and B1, remarked that his/her lack of confidence on the decision made was due to the lack of clear options among which to choose and to the lack of knowledge about the possible decisions that could be made (``\emph{There's definitely no guarantee that [this decision] will then always have a positive [outcome].}'', \han)). 

\adela, in task A1, lamented a low confidence in the decision made due to lack of confidence in the effect that a specific intervention (to perform the rehabilitation in this particular case) may have (``\emph{I don't know what I can assume [...] now that rehabilitation is false, rehabilitation has -0.15 prediction correlation, then if rehabilitation is true, the prediction correlation will be +0.85?}'' \adela). 

In other cases, however, the experts were more confident about the interpretation of plot P1 and the decision. This happened when they could easily identify the possible decisions and their effects on a trace execution exploiting their own domain knowledge. For instance, in task A1.b, \claudio felt confident to recommend no action because, according to plot P1, the trace was already predicted to have a positive outcome (``\emph{I'm more positive towards this interpretation, I feel more confident because age is nothing we have under control.}'', \claudio) .

\subsubsection{Further investigation required}
\label{sec:findings:investigate}

A common theme emerging from the interaction with both groups of experts is that often, even when declaring to be confident about the decision made, experts
still mentioned the need to do more investigation in order to fully validate the decision. 
In some cases,  experts (\manuel, \han, \niek, \anti, and \claudio) mentioned the necessity to further investigate a particular decision using additional analysis or data gathering (e.g., ``\emph{Yeah, prescribe [..] but again, I would first investigate why}'', \claudio).
\bpmml experts were more explicit in this regard mentioning specific additional data analysis techniques, requiring ``\emph{to see more data}'' (\han) 
by performing additional ``\emph{data collection, [..] full randomized trials, or bandit tests}'' (\niek).

\subsubsection{Confirming decisions combining information from different plots}
\label{sec:findings:together}

Experts showed to be able to integrate the different perspectives offered by the different plots and often used the information provided in plots shown earlier to confirm the decision made in the current task (``\emph{If I forget about the previous plot, I would say no. If I consider a combination between the two, yes}'', \pierluigi). 
In particular, \adela, \anti, \pierluigi, \claudio, \manuel, \han, and \francesco were able to spot the different perspectives shown by the plots and to infer a more complete overview of the situation presented in the decision making tasks using more than one plot at the same time (``\emph{each single plot provides you with a view of the reality.}'', \francesco). 
We can then infer that showing all the plots at the same time would have been beneficial for the comprehension of the situation at hand, ``\emph{to make more informed decision.}'' (\francesco), 
and could have even improved the experts' confidence in their decisions.

\subsubsection{Enumerate the options, providing what-if analysis}
\label{sec:findings:degrees_of_freedom}


Most of the experts (\adela, \anti, \pierluigi, \claudio, \francesco, \han, and \niek)  felt the necessity to receive additional information regarding the choices available when asked to suggest a course of action using plot P1 or P2 (
``\emph{what are my degrees of freedom?}'', \han). 
In one case (\adela), the expert had also to be reminded that taking no action was one of the possible options.

Once the possible options were enumerated, experts (\claudio, \manuel, \francesco, and \niek) also mentioned that tools supporting a prescriptive (what-if) analysis of the different options could have been useful to make more informed decisions. According to the experts, these could take the form of trace matching, i.e., matching the current trace to similar one(s) for which the outcome is known (``\emph{I don't have a counter proof. So if I had another trace in which the third event was [same value]}'', \claudio, \emph{you could do matching here and you could try to see to match this to the to the closest case [..] to see what happened to that.}'', \niek), or actionable attributes, i.e., the possibility of changing at run time the values of particular features and to see immediately the effect on the plots and the predicted outcome (\emph{we don't know what happens after this point, then you could add some actionable attribute that I can change in order to change the course of the future.}'', \francesco)). 

\subsubsection{The type of domain influences the decisions}
\label{sec:findings:domain}

The domain in which tasks are situated has clearly influenced the experts decisions (``\emph{So it's not like in the previous domain, [..] here the domain influences a lot the reasoning}'', \pierluigi). 
Specifically, for tasks in Domain A (medical), experts (\manuel, \han, \pierluigi, and \claudio) mentioned that depending on the severity of the case at hand, which was not one of the available attributes in the event log, their choice could change
 (``\emph{one might be tempted to say that the best way to make [..] is to [..]. However, it's hard to tell, because there could be some confounding attribute aspect that is not recorded in the log}'', \manuel or ``\emph{assess the situation on a case by case basis}'', \han). 
We argue that this is due to the medical domain being perceived as one where process analysis is particularly challenging because special cases are frequent and process users (physicians, nurses) often have a higher amount of freedom in deciding the course of action, e.g., on the treatments to administer \citep{MUNOZGAMA2022103994}.
The domain B (financial) was considered by experts more driven by clear and standardized procedures and therefore less challenging as far as predicting the outcome of cases is concerned. 

\subsubsection{Correlation is not causation}

The experts, in particular the \bpmml ones, mentioned multiple times that the plots that they were using were showing a correlation, e.g., between features values and predicted outcome, but that such a correlation should not be directly interpreted as a causal relation (``\emph{Yes, there is a correlation, of course, but it doesn't mean there is a causation}'' \manuel). 
%
One of them in particular (\niek) refused to carry out part of the decision making tasks for this reason (``\emph{Obviously, we do see that the fact that the patient was young contributed a lot to the fact that the positive outcome for the patient. I guess that makes sense. Young people recover more easily. That's a correlation. But I would say just from domain knowledge that's even likely to be a causation. Even though we're not sure about that. And so, that's something you can easily read in this plot. But we don't have anything to make recommendations here}'', \niek). 

Other \bpmml experts took a lighter stance in this regard, for instance, suggesting, after having mentioned the correlation vs.\ causation theme, that  ``\emph{there's definitely no guarantee}'' (\han) 
that the course of action that they suggested was going to yield the desired outcome, or that their suggested course of action was simply a ``\emph{suggestion to the domain expert}'' (\manuel). 

\subsubsection{The cost of the suggested course of action should be a decision variable}
The experts often mentioned that understanding the cost of a suggested course of action is a fundamental variable when making a decision (``\emph{I would probably suggest not to do [..] because it would incur in an additional cost}'' \manuel). 
In tasks that used plot P3 (Tasks A3 and/or B3), \adela, \anti, \pierluigi, \manuel, \han, and \niek decided to avoid making a definite decision for the process  
``\emph{depending on the cost}'' (\pierluigi), 
even if they could make a decision that was perceived as beneficial for the outcome of the process. Specifically,
\adela, \anti, \pierluigi, \manuel, \han, and \niek suggested to make the risk assessment a mandatory task for the process in task B3 of Domain B (financial) only after having considered the cost of this task. Similarly, \manuel, \han, and \niek suggested to make the rehabilitation mandatory in the process in task A3 of Domain A, only after having considered its cost.

\subsubsection{Distinguishing themes between expert groups}

Finally, we summarize here the themes that have uniquely distinguished the \bpm from the \bpmml experts in the decision making tasks.

First, we noticed that \bpmml experts tended to be more conservative in their decisions, often mentioning that the plots did not give them enough evidence to make specific recommendations and that they would rather delegate these to domain experts.   
%

Regardless the task or plot, \niek felt that the plots were not conveying enough evidence to make any specific recommendation (``\emph{I can do a little bit of diagnostics [..] but I can't make any recommendations}'', \niek). This feeling was partly echoed also by \han (``\emph{there's definitely no guarantee that [this decision] will always have a positive [outcome]}'', \han), \manuel, and \francesco,  who consistently showed to have concerns regarding any choice they made, even mentioning that they deemed necessary that someone else had the final word (e.g., requiring a ``\emph{suggestion to the domain expert}'', \manuel; 
or ``\emph{assessing the situation on a case by case basis}'', \han). 

Conversely, \bpm experts \adela, \anti, \pierluigi, and \claudio were more willing to make clear and specific recommendations based on the information available, often relying on their domain knowledge (``\emph{I have a feeling that those cases are very rare [therefore] rehabilitation should be a general practice}'' \anti)). 

The \bpm experts generally experienced a linear interaction pattern throughout the interview, answering the questions they were asked, without showing any particular additional concern regarding how the plots were produced. The questions they raised during the interviews were focused exclusively on understanding how to perform the task at hand.
For \bpm experts, we also witnessed a steep learning curve effect, which lead to the tasks on the second domain taking a considerably lower amount of time than the ones on the first domain, regardless of the order in which the domains were presented.
These experts perceived plot P3 as the easiest to understand and use. When asked, the experts reported that this was due to it being a histogram, and them being more used to interpreting histograms based on their own experience.

The \bpmml experts instead showed a deeper understanding of the proposed plots, but they also had a tendency to over-analyse the information showed in them. This, at times, hindered the flow of the interview and led their interviews to take on average longer than the ones with \bpm experts. However, the quality of the decisions made by \bpmml expert appeared not to have been affected by this tendency to over-scrutinize the information in the plots.

\subsection{RQ3: How can explanation plots be improved?}
\label{sec:findings:rq3}

Several suggestions for improving the plots have also emerged during the interviews. These revolve around three main themes, which are discussed below.


\subsubsection{Improving the interface}

Several comments we received during the interviews were directed towards improving the interface through which the plots were shown.

For example, the interface could be improved by showing all the plots together in a dashboard setting.
This would help giving the user multiple perspectives regarding the process at the same time. 
We have drawn this conclusion by analyzing the behavior of experts during the interviews. Both the \bpm and \bpmml experts used multiple plots at the same time to attain a more precise idea regarding their decisions. For instance, many experts used plot P3 to find a confirmation, at the global event log level, of the decisions they made based on the information found in P1 or P2. In alternative, experts mentioned that it would be effective to show P1 and P2 together or, at least, P2 before P1, since plot P2 represents a more general overview of the feature contributions for different prefixes, whereas P1 is a snapshot of P2 at a specific prefix length. 

In addition, a number of specific improvements have emerged regarding the elements of individual plots.  Regarding P2, the wavy lines associated with individual features were considered generally confusing and, more specifically, hinting at the visualisation of a continuous variable (``\emph{Curving is great when the domain is continuous, here is discrete, there is no 1.5 event}'', \claudio). 
In fact, the correlation values shown by P2 are discrete, because they are defined only at discrete prefix length values. Therefore, the continuous lines may be substituted by discrete plots, e.g., by markers or bars. Furthermore, P2 does not show information about the prefix length at which a feature takes on its value. For instance, feature ``Rehabilitation prescription'', in Domain A, does not have a value before the activity in which the doctor decides whether to prescribe the rehabilitation or not is executed.
This information can be helpful for decision making tasks and can be captured in the plot, for instance, by greying out the lines and/or labels associated with a particular feature until the feature takes a known value. 
Regarding P3, the green lines connecting the absolute number of traces in which the feature takes each possible value has been considered to be misleading, since they hint at a connection between the feature values, which is not meaningful for decision making (``\emph{I ended up looking at data points [...] what's the point of having a line?}'', \anti).

\subsubsection{Lack of interactive elements}
Experts have highlighted the lack of interactive elements in the plots they were asked to use.
A first level of missing interaction concerns the filtering of specific attributes, traces, or events dynamically, instead of showing one static view of them all. While, in P2, it is possible to hide the lines associated with attributes that are not deemed important for a decision (for instance, because they are associated with a constantly oscillating correlation and therefore not helpful to make decisions regarding the case outcome), a similar level of interactivity is not available in P3. Experts suggested that, in P3, it should be possible to filter only traces with certain characteristics, such as as traces similar to the one on which a decision has to be made (``\emph{In this third plot [...] I just see these numbers, should I ask what is happening behind? [...] I can divide [the data] in small groups [...] checking what it is happening behind the scene, how the data varies, how [the data] depends on the events and the variables}'', \pierluigi). 


A second level of interaction concerns supporting what-if analysis. As highlighted earlier, this is a requirement expressed in particular by the \bpmml experts, who have a more hands-on experience with predictive models and explainability methods.
What-if analysis represents a means to understand what would happen to the outcome of a process when the decision regarding a certain course of action is actually put in practice. 
Such an analysis requires, first, to be able to enumerate the decision options available in a decision making task. This should also include the option of ``doing nothing''. 

Then, the what-if analysis functionality can take two different forms. On the one hand, it could look for similar cases that have already terminated, inferring an expected outcome from them. On the other hand, it could take the form of a simulation that yields an expected outcome associated with the selected course of action.  Plots P1 and P2 are particularly suited to provide this sort of actionability, whereas this is not the case for Plot P3, which does not represent the course of actions of a single trace, but a global overview of the information available in the entire event log. 

\subsubsection{Adding information fields}

In several cases, the experts mentioned, during the interviews, the need to enrich the plots with more information. 
From a technical standpoint, experts requested more information to gauge how much they could trust the performance of the predictive model from which the plots were derived. This may involve, for all plots, showing the accuracy measures of the predictive models, like F-score or AUC.
Regarding P3, experts mentioned that showing the relative frequency of the feature values in the event log would have been more useful for supporting the decision making tasks with respect to the absolute number of traces including them. The \bpmml experts also pointed out the need to show additional statistical information in P3, such as box plots providing the confidence interval of the label value for a given feature value. 

More generally, the experts lamented the lack of information explaining the plots, such as more text describing their semantics, explanations of the acronyms used, different color codes for different types of features (e.g., to distinguish features associated with the occurrence of activities in a trace from other features) and a more clear explanations of the term ``correlation'' in plots P1 and P2.

\section{Discussion and Conclusion}
\label{sec:disc}
In this section, we analyze and discuss the results obtained from the user study by drawing the main implications for theory (Section~\ref{ssec:theory}), for practice (Section~\ref{ssec:practice}), as well as by highlighting the limitations (Section~\ref{ssec:limitations}) of the evaluation carried out. 

\subsection{Implications for theory}
\label{ssec:theory}

The results presented in the previous section have highlighted that the explanations plots need to be complemented by tools supporting the what-if analysis to foster the experts' confidence in the decisions made. Future research should investigate the most effective way of implementing such tools, e.g., whether focusing on finding similarities between historical cases and the one for which a decision should be made, or implementing full-fledged simulations tools. Reinforcement learning~\citep{DBLP:books/lib/SuttonB98} and bandit algorithms~ \citep{DBLP:journals/ftml/BubeckC12,lattimore2020bandit} may also be considered to allow decision support tools to learn the best course of action (policy) that maximizes an expected gain. 

Future research should also address the issue of guaranteeing that the explanation plots provided to experts convey the information required to establish causality, between feature values and process outcomes in the case of this paper, rather than simply correlation. In this direction, anchors~\citep{DBLP:conf/aaai/Ribeiro0G18} are a post-hoc explanation technique that yields if-then rules explaining the behavior of the underlying model, together with an indication of the precision and coverage of the rules. As far as we are aware, an approach customizing anchors to the case of PPM is missing in the literature.

From a human factor standpoint, more research is needed to understand which skills are required by decision makers to perform well in the decision making tasks with the support of explanation plots. Counter-intuitively, our analysis has highlighted that experts lacking deep ML knowledge felt more comfortable using the plots. Future research should investigate to what extent other variables such as decision making experience, process modeling expertise, or process domain knowledge may influence the ability to make correct decisions in PPM scenarios.
Along this line, the opportunities for improving the current plots that we have been identified in this work should be validated with different types of decision makers and in different classes of decision making tasks.

Finally, future research should investigate to what extent the vision of self-optimizing business processes can be implemented with the support of AI tools within the business process lifecycle, i.e., whether the decision making tasks, possibly exploiting the feedback produced by explainable AI techniques, can be fully delegated to automated tools, bypassing human decision makers. In this direction, recent research on prescriptive monitoring of business processes~\citep{bozorgi2021prescriptive,weinzierl2020prescriptive} has dealt with (semi-automatically) incorporating the effects of what occurs during the execution of a business process (for instance as the result of human actions) while predicting the value of aspects of interest in the long term.



\subsection{Implications for practice}
\label{ssec:practice}
Our findings revealed how different individuals utilize explanation plots for decision making and can thus serve as a basis for suggestions on how to employ them in the context of PPM.

While individuals appreciated the plots as a valuable source of information, we also found that they often did not perceive the information presented to be rich enough to make a decision confidently. When employing such plots, it is thus important to be able to provide additional information to decision makers. Our findings are in line with those of \citep{DBLP:conf/chi/LimDA09} who proved that employing explanation plots supports understanding.

One additional piece of information that the participants of our study deemed to be particularly relevant was the causality of the presented predictions. They did not perceive showing correlations to be sufficient to make a decision especially in a medical context. Providing causality-related information can thus be perceived to be important when using explanation plots for decision making in the context of PPM. This need has also been reported by \citep{lipton2001good} who found causality to be one of the four key features fostering explanation.

Together with information about causality, participants also asked for domain-related information, for details about the specific case that were not present in the explanation plots, for options to filter the information that was available in the plots, and for the possibility to perform additional analysis like on-the-fly what-if analysis and statistical significance tests.Domain-related information and case details can be easily added, as well as suitable filters. The additional analysis suggested by the participants might be more difficult to provide. The need for a what-if analysis has also been reported in~\citep{davis1994user}. However, they also found that providing such analysis can create an `illusion of control' that causes the users to overestimate its effectiveness. When providing more information, it is thus necessary to carefully consider which information to provide and to whom.

We also found differences between the expectations of \bpm experts and \bpmml experts regarding their need for information to make a decision.
For \bpm experts, 
we suggest to pay particular attention to the wording of the plot labels and captions. For instance, in our study, the \bpm experts struggled with the term `Correlation' as it is too vague if not well explained and contextualized. They also had issues in deciding which actions to take and asked which actions would even be possible. To support their decision process, we would thus suggest to provide examples of possible actions that were taken in past process executions similar to the one of interest as explored in \citep{DBLP:conf/er/LuS07}.
Finally, the \bpm experts also assumed that the information shown was related to statistics on completed process executions and they did not immediately realized that the plots serve as additional information complementing the prediction tasks. We would thus propose to stress the goal of the plots by also improving their captions.

On the other hand, \bpmml experts asked for information regarding predictions and the way they were computed including the data distribution of the original data and measures representing the prediction accuracy of the predictive model. This information can be easily provided, and would make the decision making process more informed. We also observed that the \bpmml experts avoided reading the text in the captions. So we would make these elements of the interface more prominent or use pop-ups when the user interacts with a given element. Finally, the \bpmml experts suffer from the information overload coming from their expertise often resulting in getting lost in the details of the plots, and the descriptions~\citep{buchanan2001information}. We therefore suggest to complement these plots with plots that provide information at a higher level of abstraction that focus more on the business value of the information provided and on the decision that needs to be made rather than on the model-specific information provided by the plots analyzed in this user evaluation.

\subsection{Limitations}
\label{ssec:limitations}
The goal of our study was to investigate how users make sense of explanation plots and how they use them for decision making. Furthermore, we aimed at exploring how explanation plots can be improved. It is thus reasonable to conduct a qualitative observational study~\citep{lazar2017research}. There are, however, innate limitations associated with this study design. We interviewed individuals from different backgrounds, domain expertise and expertise related to BPM and ML. Despite making a reasonable selection of the participants, it is not possible to generalize findings beyond our study context since studying different individuals with different backgrounds from different domains and different levels of expertise might yield different results. Moreover, the study was conducted by a team of researchers which poses a threat to validity since different researchers might perceive the reactions of study participants differently. To minimize this threat, we ensured that throughout the process of the study, which included the preparation of the study materials, the conduction of the study and the analysis of the study results, at least two individuals from the research team collaborated on each step to avoid depending on the perception of individual researchers. We also opted for studying a specific artificial setting utilizing specific plots and asking predefined questions. This can lead to observations and interpretations that might not have happened or might have happened in a different way in a real-life setting. To mitigate this threat, we made a state-of-the-art founded selection of the explanation plots, study domains and decision tasks. We acknowledge though that there is a remaining risk associated with studying an artificial rather than a real setting. We are willing to accept this threat because it allowed us to compare study findings across subjects which would not be possible when studying a real case. Finally, we abstain from making causal claims providing instead a rich description of the observed behavior and reported perceptions of the study participants based on which we discuss differences in how different individuals made sense of explanation plots and used them for decision making.

\appendix

\section{Appendix}
\label{sec:appendix:figures}
This appendix contains the study material used by the investigators and includes the different explanation plots, an introduction to each plot, the questions that the investigators asked, the expected answers and the interaction capabilities of the interface.

\subsection{Plots and descriptions}

\begin{figure}[!h]
    \centering
    \includegraphics[width=.9\linewidth]{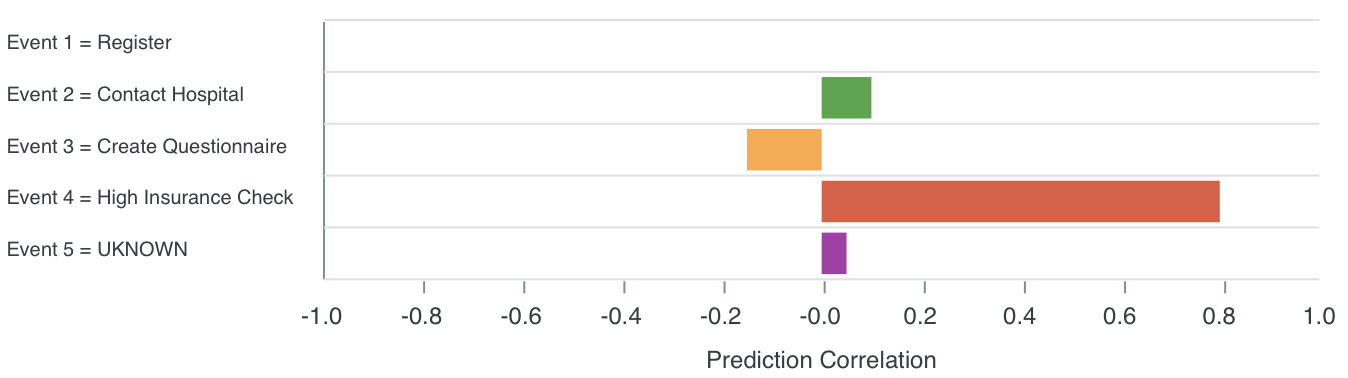}
    \caption{Plot P1 for the comprehension task.}
    \label{fig:appendix:plotp1}
\end{figure}

\begin{figure}[!h]
    \centering
    \includegraphics[width=0.8\linewidth]{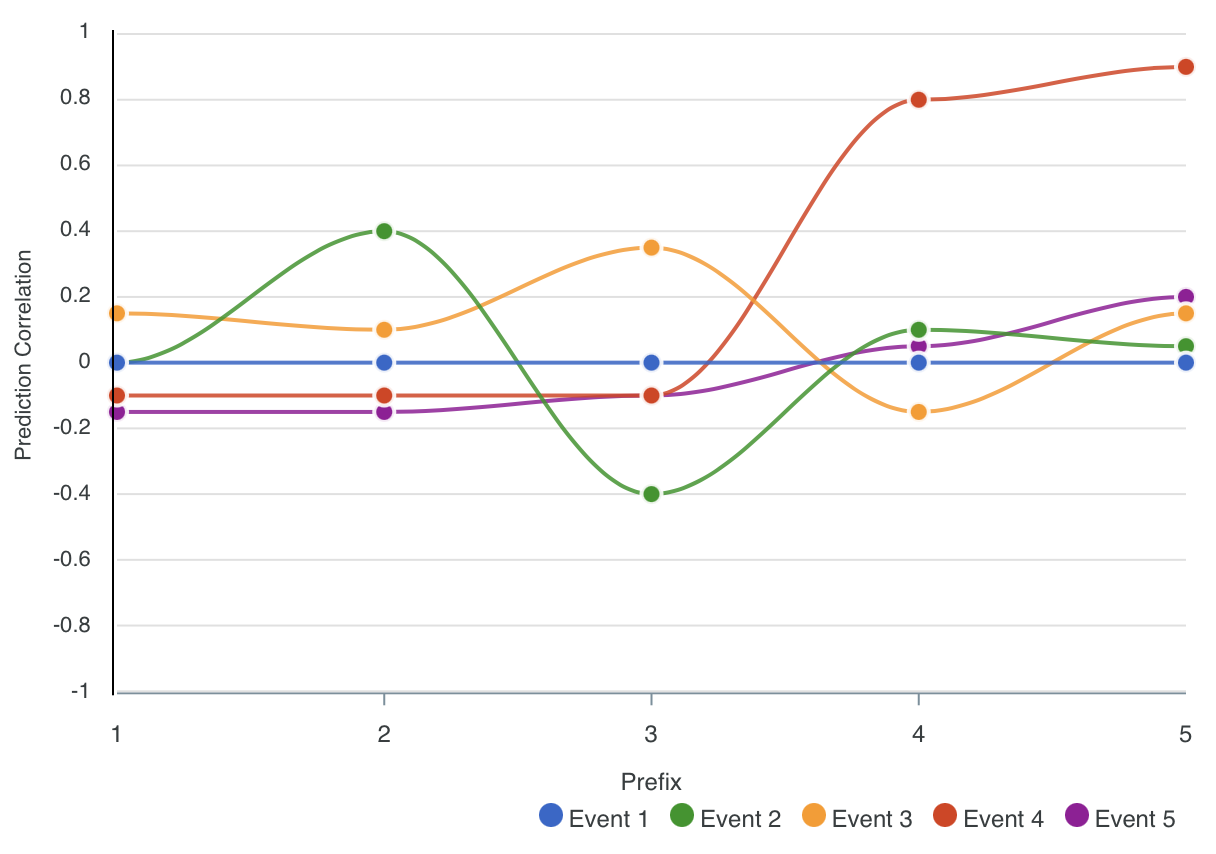}
    \caption{Plot P2 for the comprehension task.}
    \label{fig:appendix:plotp2}
\end{figure}

\begin{figure}[!h]
    \centering
    \includegraphics[width=0.9\linewidth]{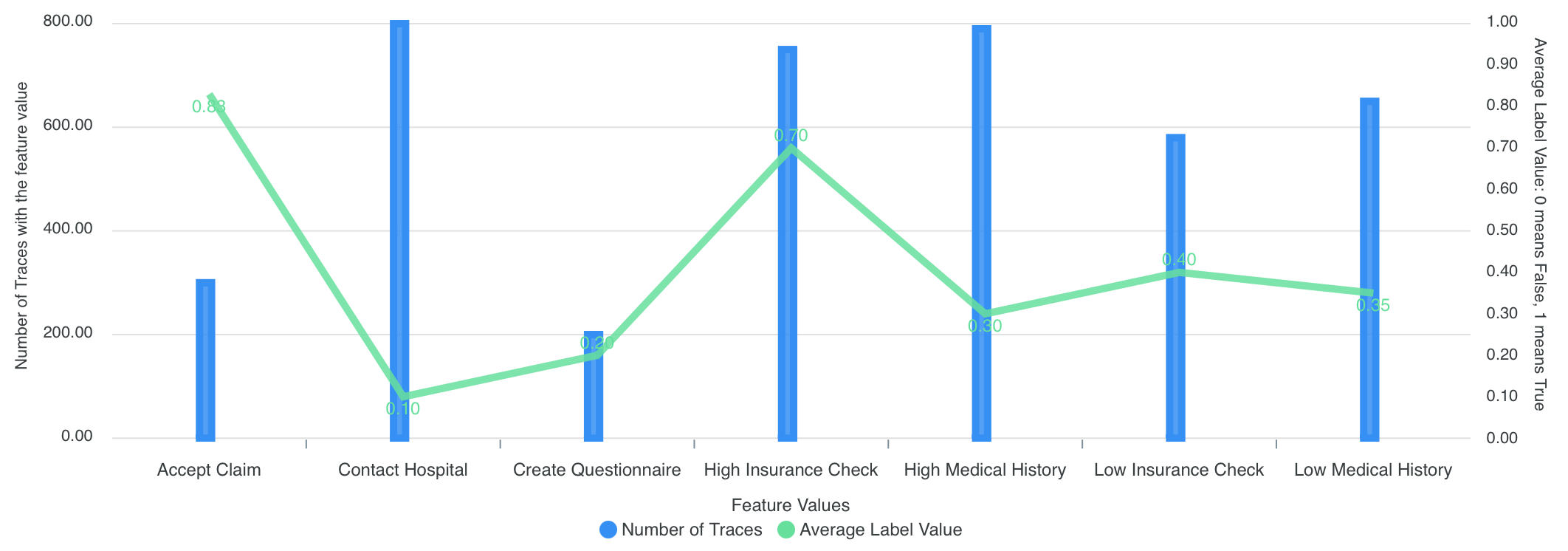}
    \caption{Plot P3 for the comprehension task.}
    \label{fig:appendix:plotp3}
\end{figure}

Let us assume to have an insurance claim event log that collects claim requests. We use this log to train a predictive model and predict whether the claim will be accepted (TRUE) or rejected (FALSE).
Let us assume that the log is encoded using the simple-index encoding (an example of this encoding is shown to the participant) for training the predictive model.
Let us consider the following explanation plots: the first two plots (Figures~\ref{fig:appendix:plotp1} and~\ref{fig:appendix:plotp2}) refer to trace 4633, while the third plot (Figure~\ref{fig:appendix:plotp3}) refers to a set of incomplete process instances.

Questions asked by the investigator:
\begin{itemize}
    \item What do the different plots show? 
    \item Which are the feature(s) influencing most the prediction related to trace 4633? Why? 
    \item How confident are you about your interpretation?
    \item Which features are more stable over time with respect to their importance towards the acceptance prediction for trace 4633? Why?
    \item What is the value of the feature \emph{Event 4} influencing most the acceptance prediction over all the traces in the log? Why?
    \item When thinking about which answer to give, which information did you miss?
\end{itemize}

\subsection{Two domains}
\label{sec:appendix:domains}

\subsubsection{Domain A}
Let us assume to have an event log pertaining to the treatment of patients with fractures. Every process instance in the log starts with activity \act{Examine patient}. If activity \act{Perform X-ray} is performed, then \act{Check X-ray risk} must be performed before it, without other executions of \act{Perform X-ray} in between. Activities \act{Perform reposition}, \act{Apply cast} and \act{Perform surgery} 
require that \act{Perform X-ray} and \act{Make Prescriptions} are executed before they are executed. If \act{Perform surgery} is performed, then \act{Prescribe rehabilitation} is performed eventually after it. Finally, after every execution of \act{Apply cast}, eventually \act{Remove cast} is executed and, viceversa, before every execution of \act{Remove cast}, \act{Apply cast} must be performed. Each process instance refers to a patient and her age is also available in the event log. Moreover, the activity \act{Make Prescriptions} includes the event attributes \act{treatment type} and \act{rehabilitation prescription}, which record the type of treatment prescribed and whether or not the patient should carry on the rehabilitation.

We aim at predicting whether the patient will recover quickly (TRUE) or not (FALSE).

\begin{figure}[!h]
    \centering
    \includegraphics[width=0.6\linewidth]{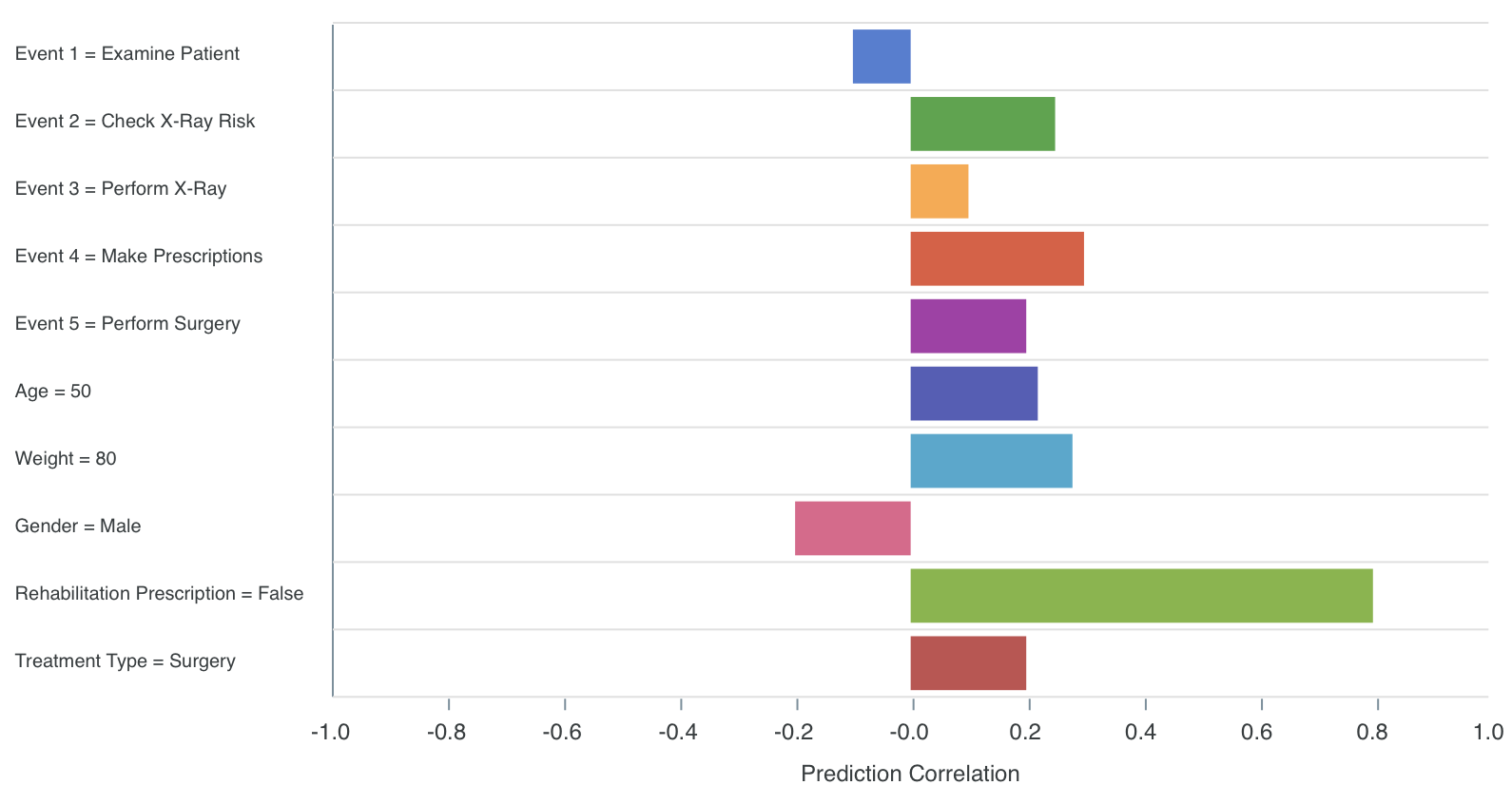}
    \caption{Plot P1 for the decision making task A1.a.}
    \label{fig:appendix:plotA1a}
\end{figure}

\paragraph{Task A1.a}
Let us consider an incomplete trace of a patient who has carried out one of the treatments, i.e., \act{Perform Reposition}, \act{Apply cast} or \act{Perform Surgery}. For this patient, the prediction is that she will not recover soon from the fracture. The explanation of the prediction is reported in Figure~\ref{fig:appendix:plotA1a}.

Questions asked by the investigator (Q) and projected answers based on the information provided (A):
\begin{question_answer}
    \item How do you interpret the plot? 
    \itema The patient does not recover because rehabilitation has not been prescribed.
    \item Based on the information presented in the plot, what course of action would you suggest?
    \item Would you recommend carrying on the rehabilitation? Why?
    \itema Yes, because the negative outcome, i.e., the fact that the patient will not recover soon, is due to not carrying out the rehabilitation.
\end{question_answer}

\begin{figure}[!h]
    \centering
    \includegraphics[width=0.6\linewidth]{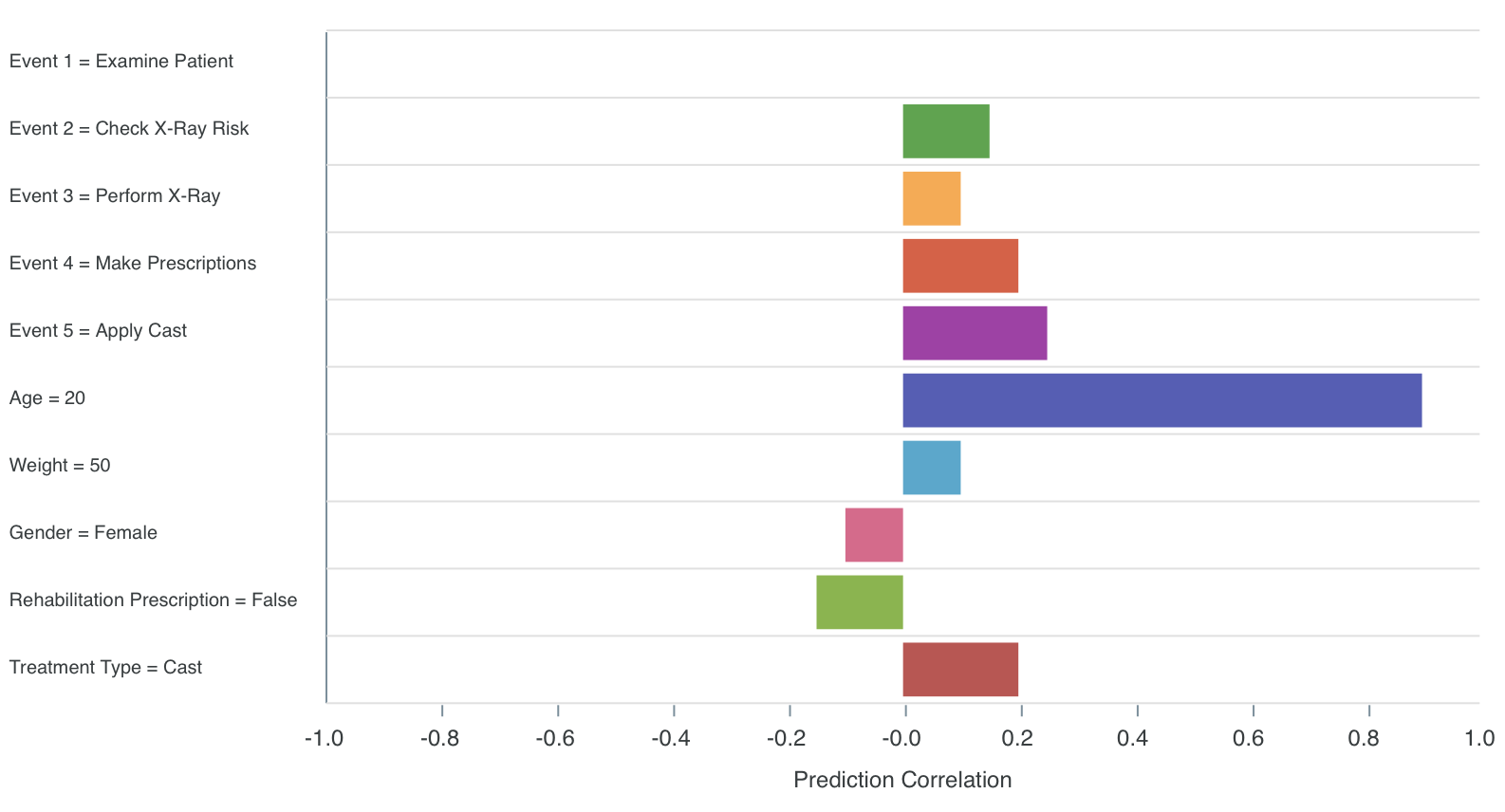}
    \caption{Plot P1 for the decision making task A1.b.}
    \label{fig:appendix:plotA1b}
\end{figure}

\paragraph{Task A1.b}
Let us now consider the incomplete trace of another patient who has carried out one of the treatments, i.e., \act{Perform Reposition}, \act{Apply cast} or \act{Perform Surgery}.
For this patient, the prediction is that she will recover soon from the fracture. The explanation of the prediction is reported in Figure~\ref{fig:appendix:plotA1b}.

Questions asked by the investigator (Q) and projected answers based on the information provided (A):
\begin{question_answer}
    \item How do you interpret the plot? 
    \itema The patient will recover because of her age.
    \item Based on the information presented in the plot, what course of action would you suggest?
    \item Would you recommend carrying on the rehabilitation? Why? 
    \itema No, because the positive outcome, i.e., the fact that the patient will recover soon is due to the age of the patient.
\end{question_answer}

\begin{figure}[!h]
    \centering
    \includegraphics[width=.9\linewidth]{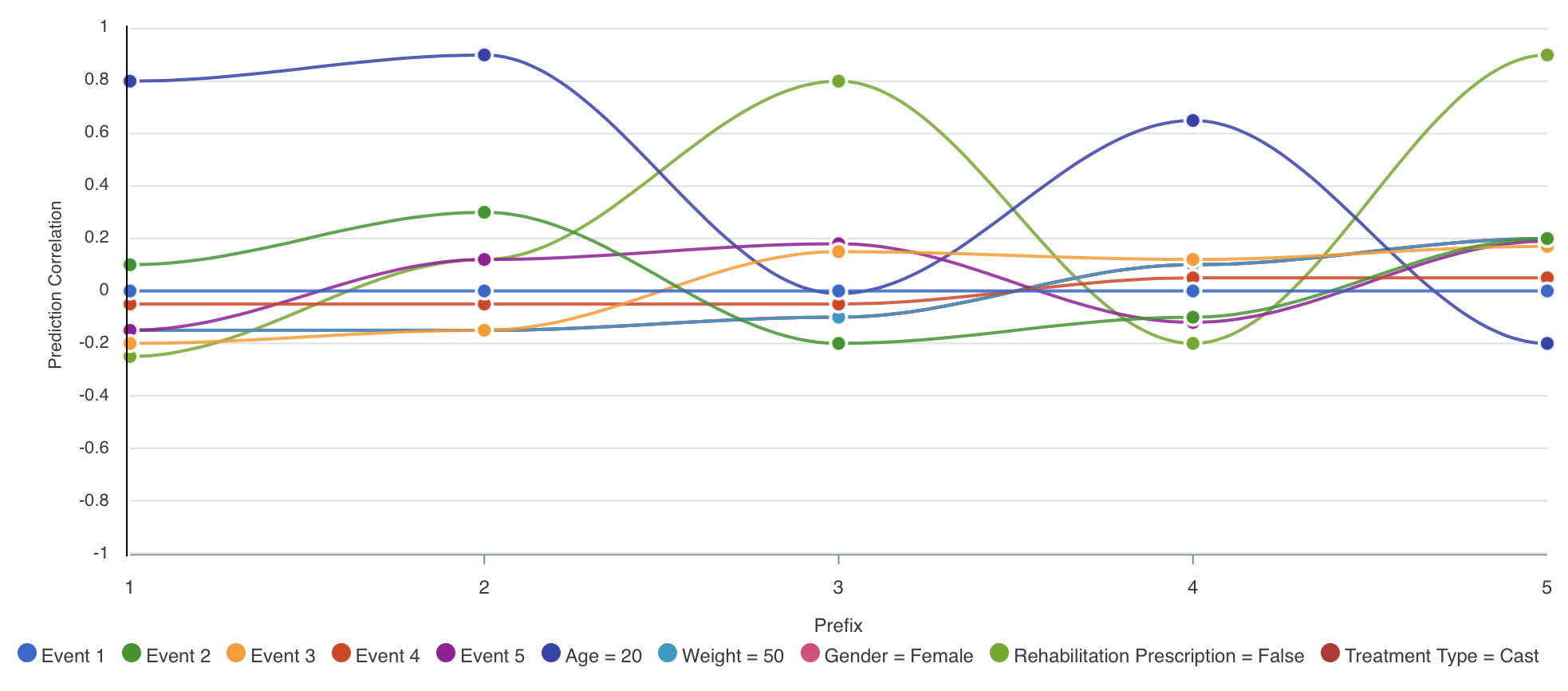}
    \caption{Plot P2 for the decision making task A2.}
    \label{fig:appendix:plotA2}
\end{figure}

\paragraph{Task A2}
Let us consider an incomplete trace of a patient who has carried out one of the treatments, i.e., \act{Perform Reposition}, \act{Apply cast} or \act{Perform Surgery}.
For this patient, the prediction is that she will recover soon from the fracture. The explanation of the prediction is reported in Figure~\ref{fig:appendix:plotA2}.

Questions asked by the investigator (Q) and projected answers based on the information provided (A):
\begin{question_answer}
    \item How do you interpret the plot? 
    \itema The explanation is initially (when looking only at the first two events) that this is due to the age, then (when looking at the first three events) that this is due to the rehabilitation prescription, then  (when looking at the first four events) that this is due again to the age and, finally (when looking at the first five events) that this is due to the rehabilitation prescription.
    \item Based on the information presented in the plot, what course of action would you suggest?
    \item Would you recommend carrying on the rehabilitation? Why? 
    \itema No, because the importance of the rehabilitation prescription is rather unstable.
\end{question_answer}

\begin{figure}[!h]
    \centering
    \includegraphics[width=.7\linewidth]{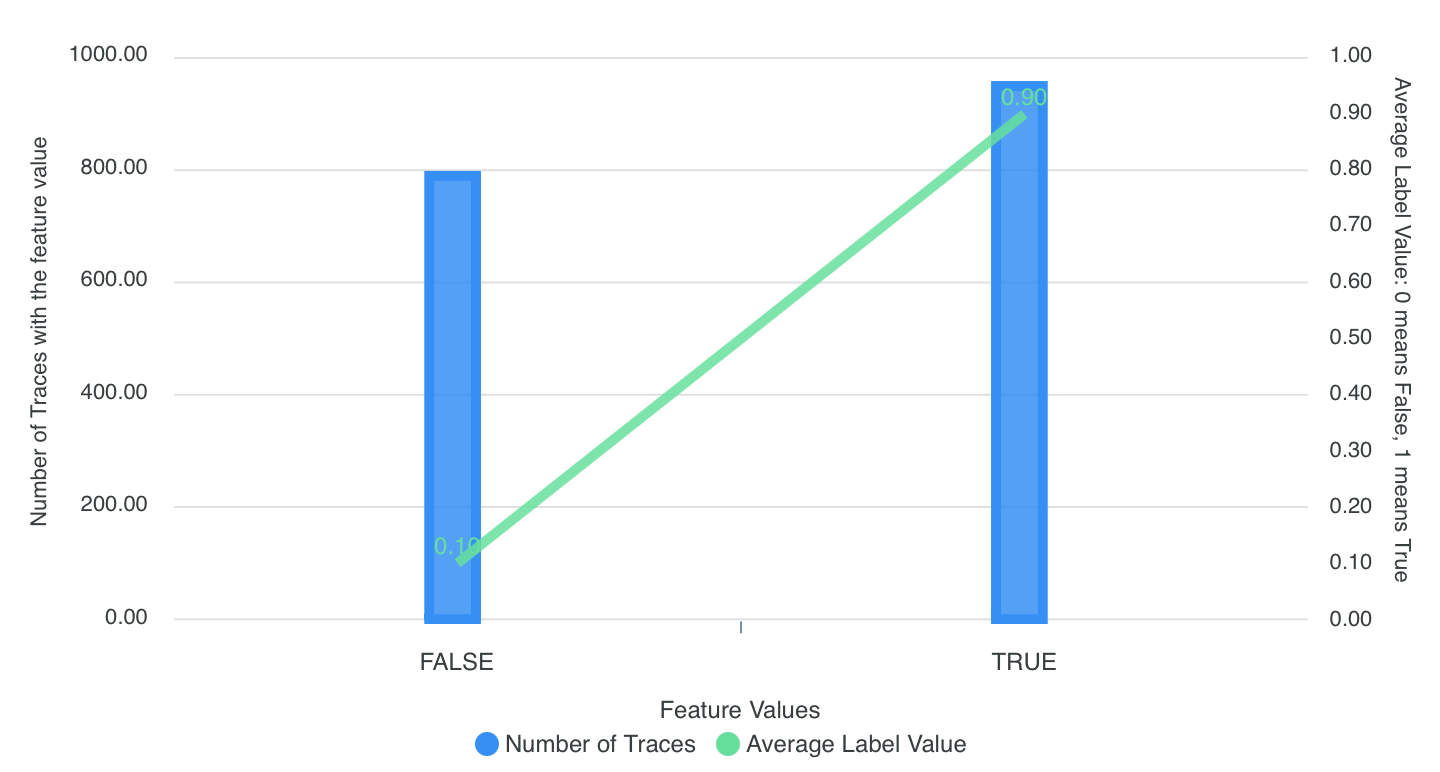}
    \caption{Plot P3 for the decision making task A3.}
    \label{fig:appendix:plotA3}
\end{figure}

\paragraph{Task A3}
Let us consider a set of process executions related to patients who have carried out one of the treatments, i.e., \act{Perform Reposition}, \act{Apply cast} or \act{Perform Surgery}.
For some of these patients, the prediction is that they will recover soon from the fracture, for others, the prediction is that it will take time for them to recover. The explanation of the predictions is reported in Figure~\ref{fig:appendix:plotA3}.

Questions asked by the investigator (Q) and projected answers based on the information provided (A):
\begin{question_answer}
    \item How do you interpret the plot? 
    \itema The explanation is that when the rehabilitation is prescribed, the prediction goes towards a fast recovery, when the rehabilitation is not prescribed the recovery is slow.
    \item Based on the information presented in the plot, what course of action would you suggest?
    \item Would you recommend rehabilitation as a general practice for all patients? 
    \itema Yes, because cases in which the rehabilitation was performed are usually experiencing a fast recovery.
\end{question_answer}

\subsubsection{Domain B}
Let us consider a process at a bank to handle the closing of a bank account by the finance operations department. A request is first created, by either the owner of the account or a 3rd-party, such as their attorney. Then, the request is evaluated. As part of the evaluation, a \act{risk assessment} may also occur, which involves checking for abnormal transactions in the account history or inquiring with the state fiscal authority to understand whether the account has been involved in illicit or suspicious activities. \act{Risk assessment} is optional and may also be executed later in the process.  Then, the outstanding balance of the account is determined, including pending payments; before the closing request can be filed, an investigation of the account owner’s heirs has to be executed in order to understand to whom the outstanding balance should be transferred. After a final evaluation, the outcome of the case is determined and the process terminates. The outcome of the process can be either of the following: the request will be executed (TRUE) or the request will be sent to the credit collection (FALSE) for a more thorough assessment. The latter may happen, for instance, when there are irregularities in the request, pending payments, or the heirs are unreachable.  Since credit collection takes long and involves a high amount of resources, the bank would like to minimize the number of requests sent to credit collection. 

\begin{figure}[!h]
    \centering
    \includegraphics[width=.6\linewidth]{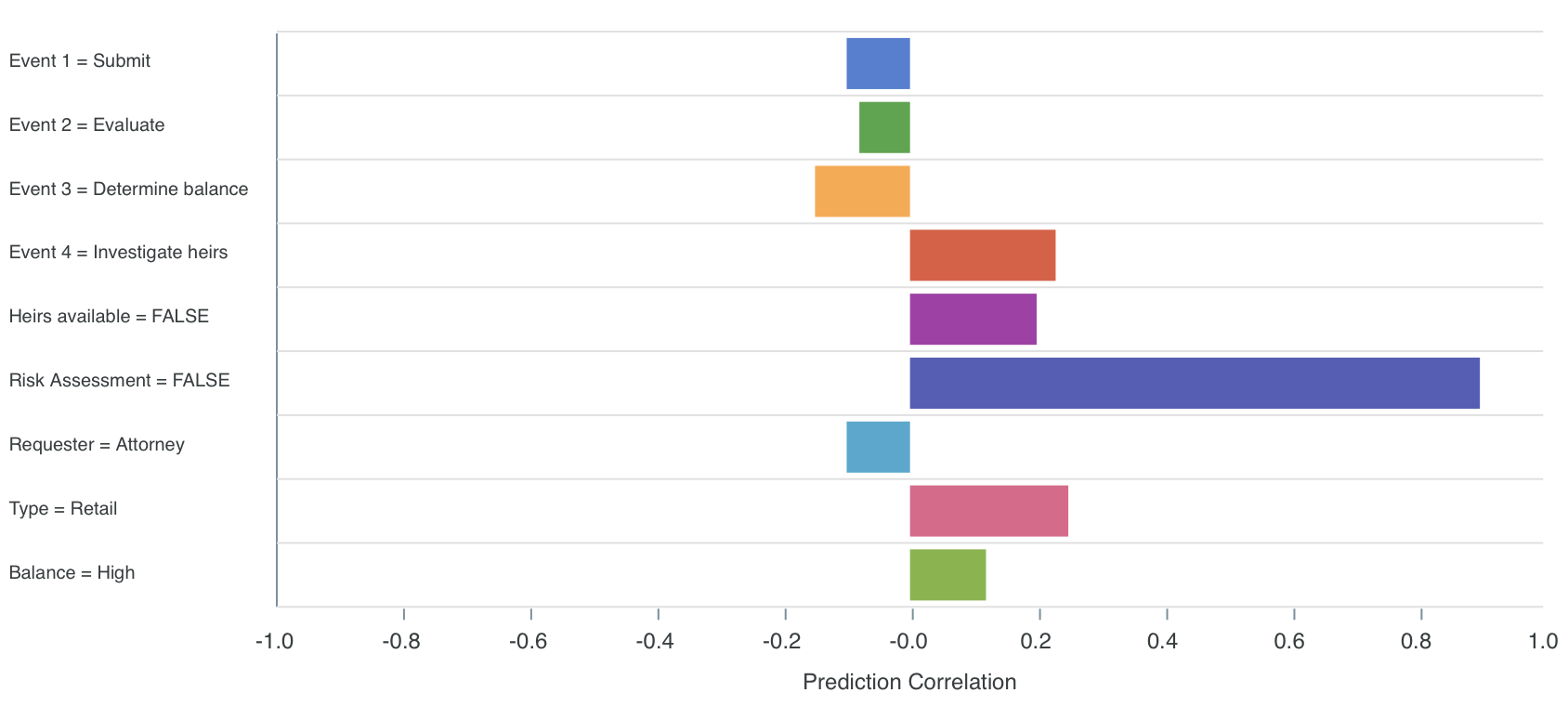}
    \caption{Plot P1 for the decision making task B1.}
    \label{fig:appendix:plotB1}
\end{figure}

\paragraph{Task B1}
Let us consider an incomplete trace (trace 2) of a bank account closure. The label of this trace is FALSE (meaning that it is sent to credit collection). The explanation of this prediction is shown in Figure~\ref{fig:appendix:plotB1}.

Questions asked by the investigator (Q) and projected answers based on the information provided (A):
\begin{question_answer}
    \item How do you interpret the plot? 
    \itema The request will be sent to credit collection because a risk assessment was not executed.
    \item Based on the information presented in the plot, what course of action would you suggest?
    \item Would you recommend running the risk assessment before the end of the case? Why? 
    \itema Yes, the user should suggest to run the risk assessment because the feature risk assessment = false is highly correlated to the negative outcome.
\end{question_answer}

\begin{figure}[!h]
    \centering
    \includegraphics[width=.9\linewidth]{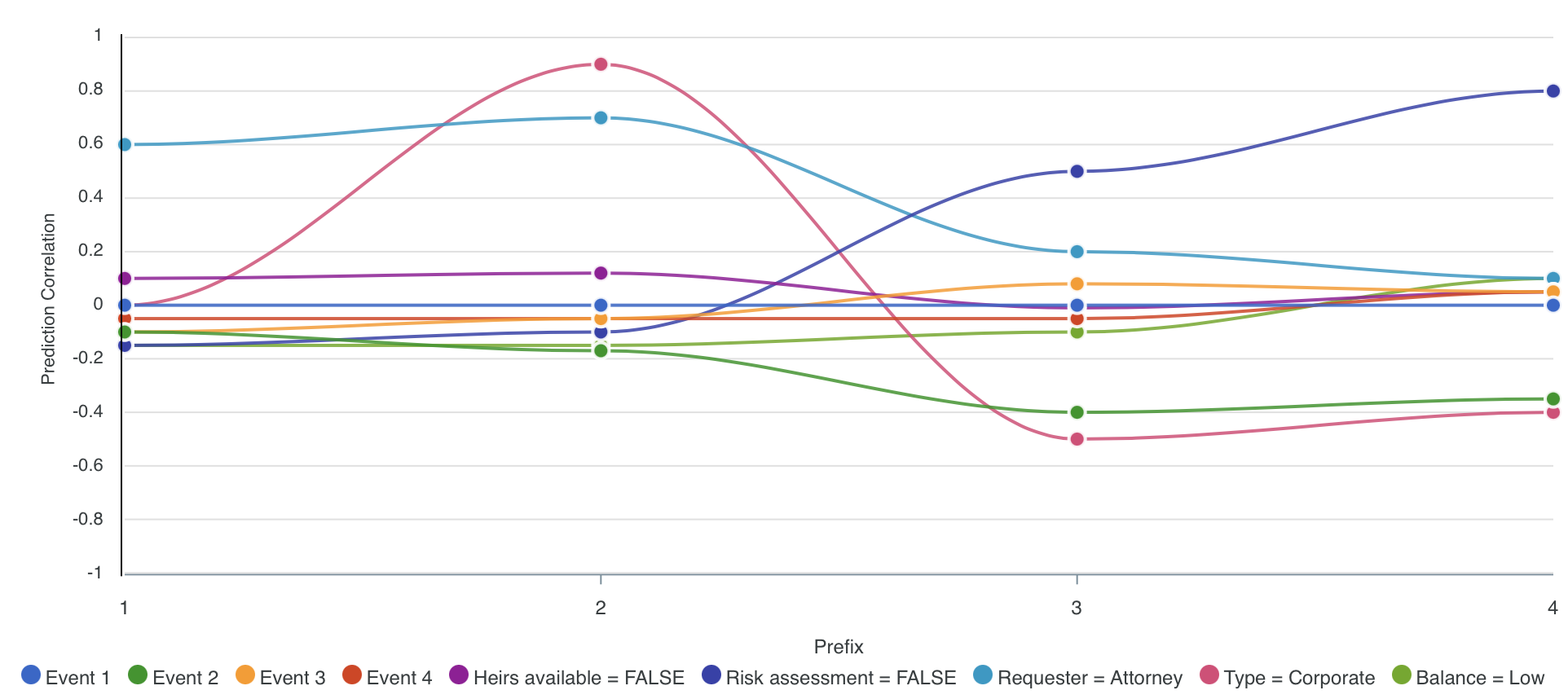}
    \caption{Plot P2 for the decision making task B2.}
    \label{fig:appendix:plotB2}
\end{figure}

\paragraph{Task B2}
Let us consider trace 3 at prefix 5. The label of this trace is FALSE (meaning that it is sent to credit collection). The explanation plot is shown in Figure~\ref{fig:appendix:plotB2}.

\begin{figure}[!h]
    \centering
    \includegraphics[width=.7\linewidth]{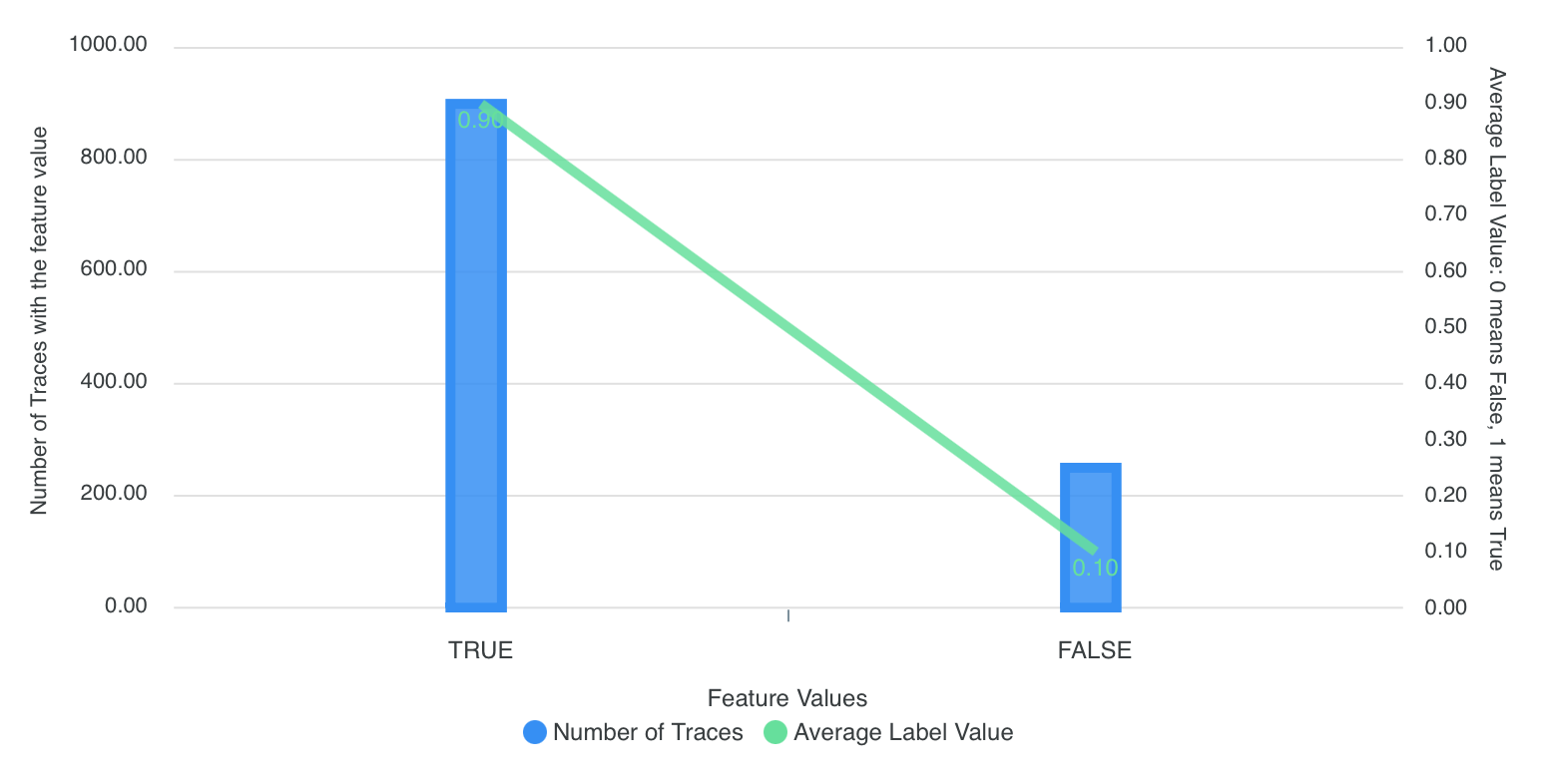}
    \caption{Plot P3 for the decision making task B3.}
    \label{fig:appendix:plotB3}
\end{figure}

Questions asked by the investigator (Q) and projected answers based on the information provided (A):
\begin{question_answer}
    \item How do you interpret the plot? 
    \itema Initially, it seems like risk assessment is not necessary because the false label is explained by the type of requester (attorney) and the type of account (corporate), but then it becomes clear that the false label is explained by the absence of risk assessment.
    \item Based on the information presented in the plot, what course of action would you suggest?
    \item Would you recommend running the risk assessment before the end of the case? Why? 
    \itema Yes, from prefix 3 onwards it becomes clear that the false label is explained by the absence of risk assessment.
\end{question_answer}

\paragraph{Task B3}
Let us consider a set of process executions and let us focus on the feature capturing whether the risk assessment was executed or not. Depending on the value of this feature, for some requests, the prediction is that they can be executed, for others, the prediction is that they will be sent to credit collection. The explanation is shown in Figure~\ref{fig:appendix:plotB3}.

\begin{figure*}[!t]
    \centering
    \begin{subfigure}[b]{0.45\textwidth}
        \centering
        \includegraphics[width=\linewidth]{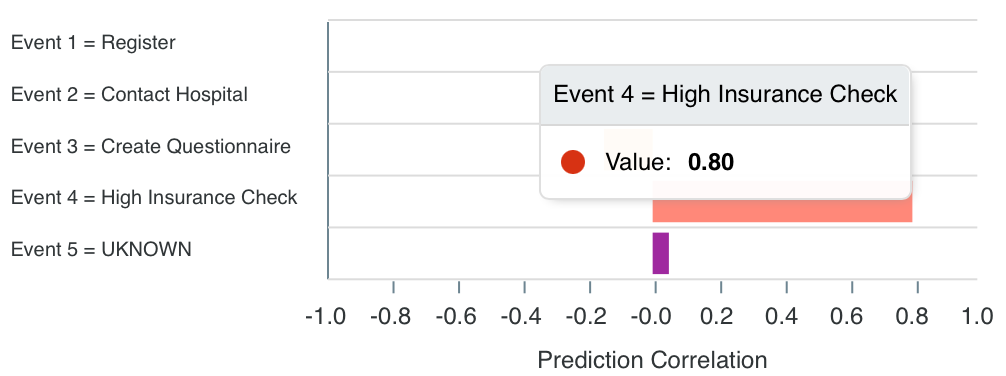}
        \caption{\small Detailed view the prediction correlation of Event 4.}
        \label{fig:appendix:hoverP1}
    \end{subfigure}
    \hfill
    \begin{subfigure}[b]{0.5\textwidth}  
        \centering
        \includegraphics[width=\linewidth]{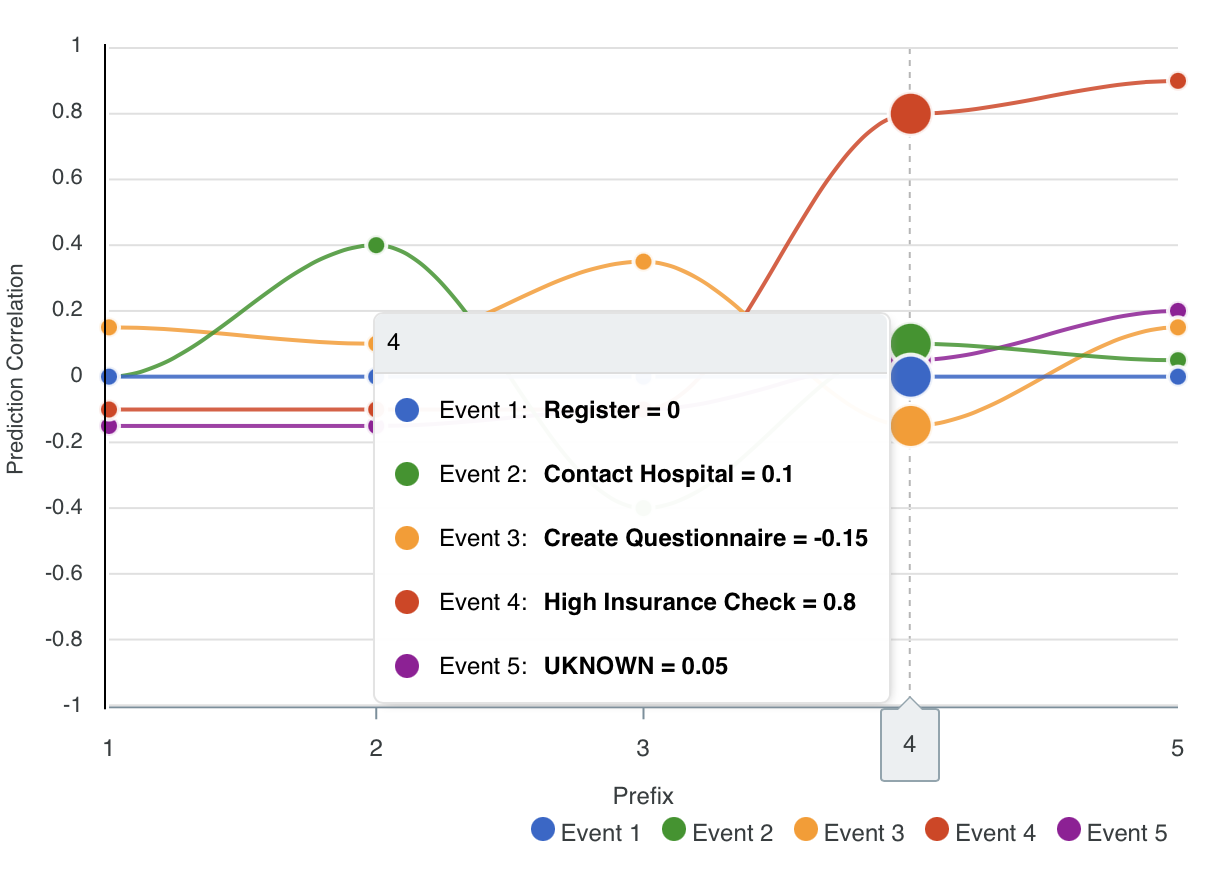}
        \caption{\small Detailed view showing the prediction correlation of all features when Event 4 occurs. }
        \label{fig:appendix:hoverP2}
    \end{subfigure}
    \vskip\baselineskip
    \begin{subfigure}[b]{\textwidth}  
        \centering
        \includegraphics[width=.9\linewidth]{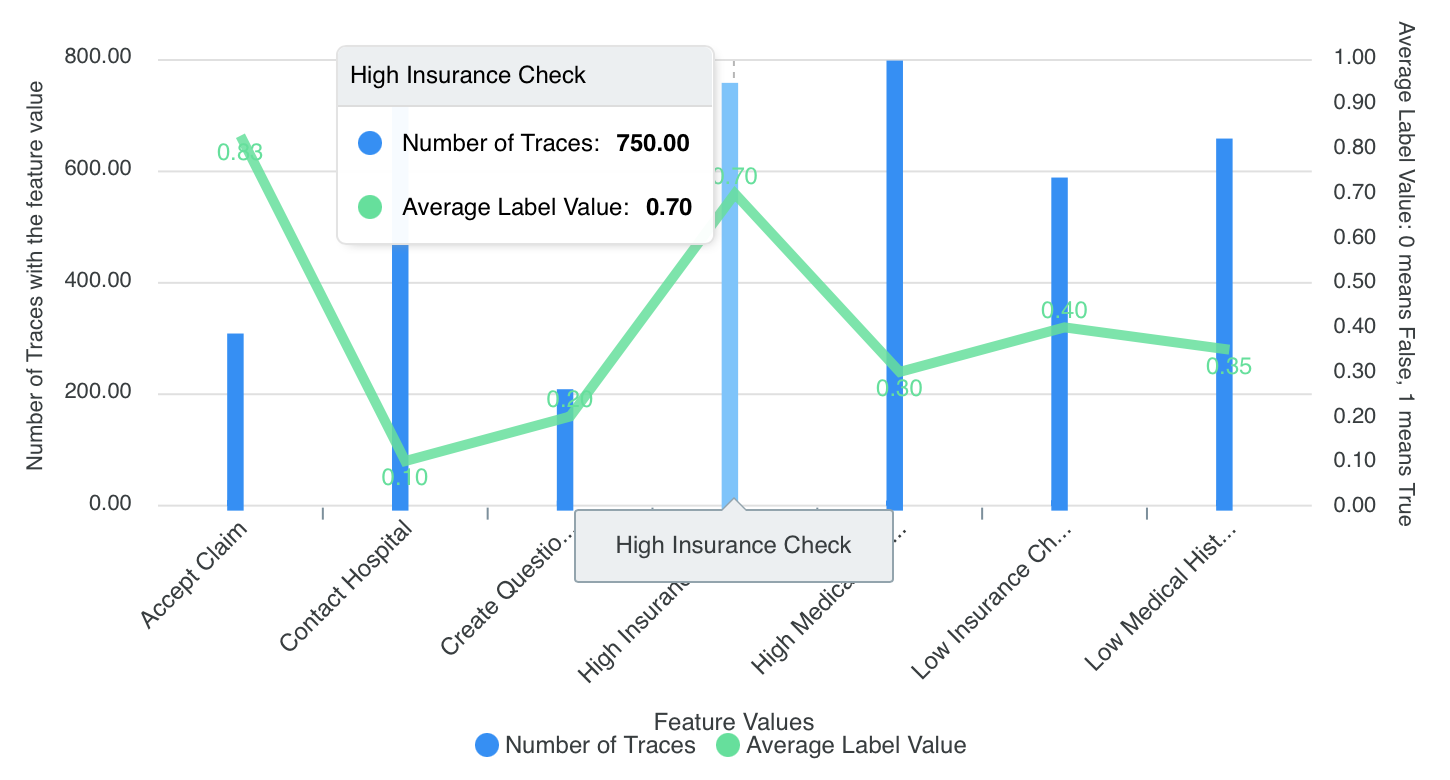}
        \caption{\small Detailed view for Plot P3 showing the number of traces containing the feature value 'High Insurance Check' and the average label value for this subset of traces.}
        \label{fig:appendix:hoverP3}
    \end{subfigure}
    \caption{\small Examples of detailed views for Plot P1, Plot P2, and Plot P3.} 
    \label{fig:appendix:hover}
\end{figure*}

\begin{figure}[!h]
    \centering
    \includegraphics[width=.6\linewidth]{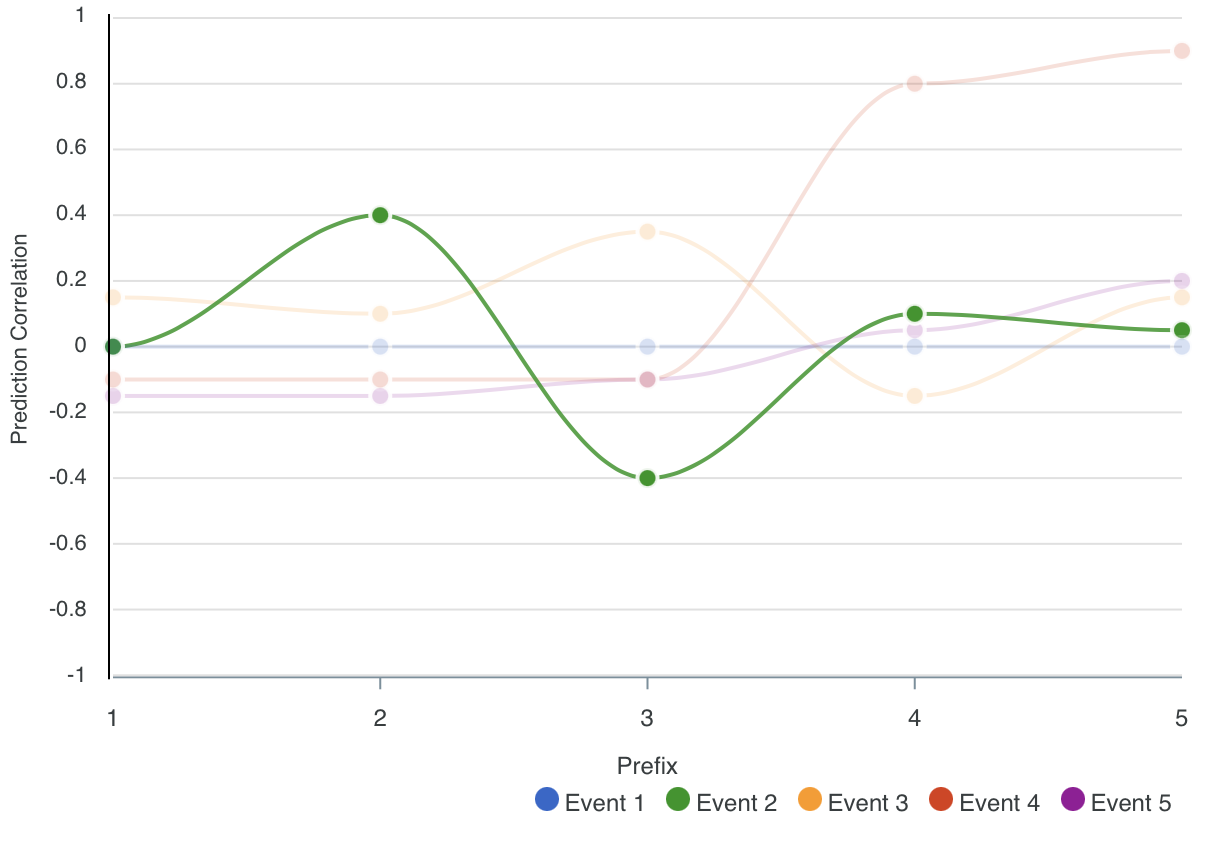}
    \caption{Focused view retaining feature Event 2 and blurring out the others.}
    \label{fig:appendix:focusMode}
\end{figure}

\begin{figure*}[!h]
    \centering
    \begin{subfigure}[b]{0.475\textwidth}
        \centering
        \includegraphics[width=\textwidth]{figures/comprehension/plot2.png}
        \caption{{\small Unfiltered view with five features.}}   
        \label{fig:appendix:filterModeA}
    \end{subfigure}
    \hfill
    \begin{subfigure}[b]{0.475\textwidth}  
        \centering 
        \includegraphics[width=\textwidth]{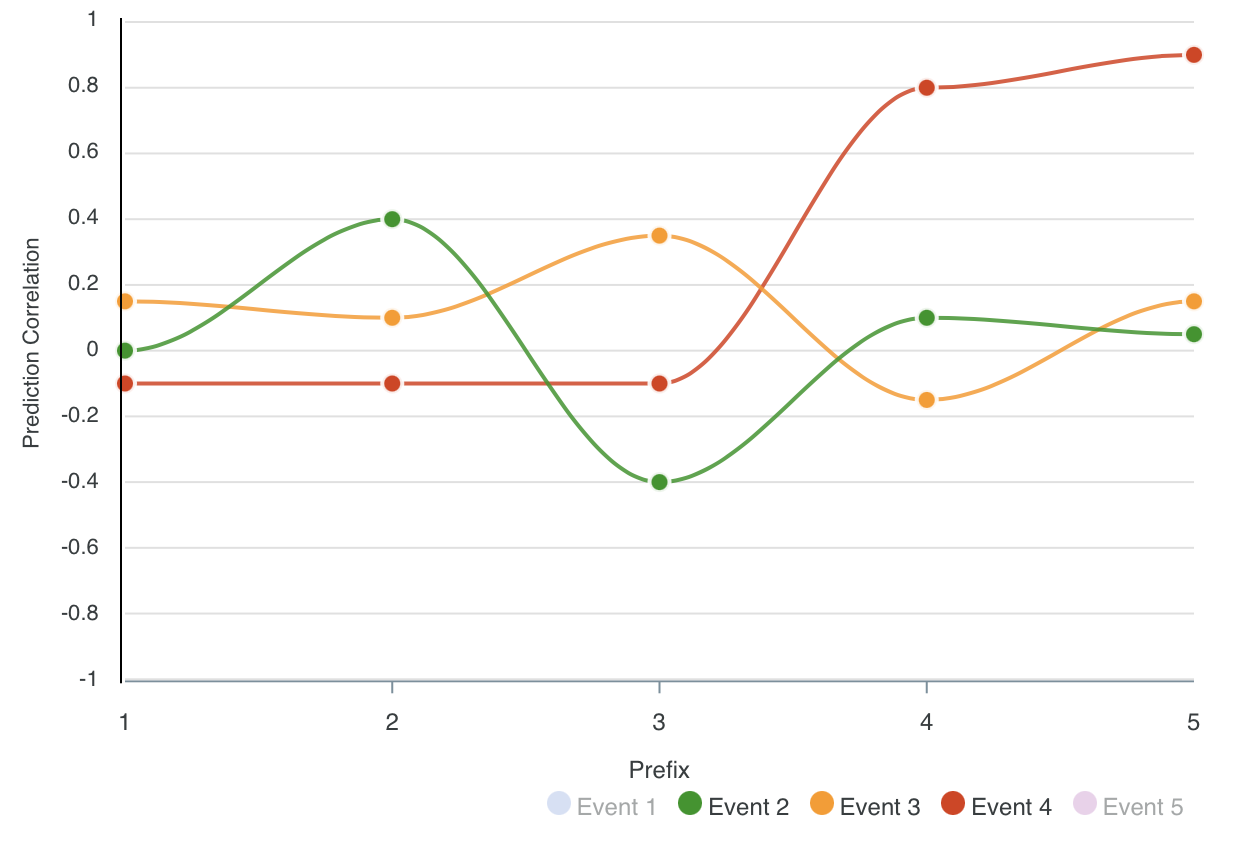}
        \caption{{\small Filtered view retaining only features Event 2, Event 3, and Event 4.}}    
        \label{fig:appendix:filterModeB}
    \end{subfigure}
    \vskip\baselineskip
    \begin{subfigure}[b]{0.475\textwidth}   
        \centering 
        \includegraphics[width=\textwidth]{figures/plotA2.png}
        \caption{{\small Unfiltered view with ten features.}}    
        \label{fig:appendix:filterModeC}
    \end{subfigure}
    \hfill
    \begin{subfigure}[b]{0.475\textwidth}   
        \centering 
        \includegraphics[width=\textwidth]{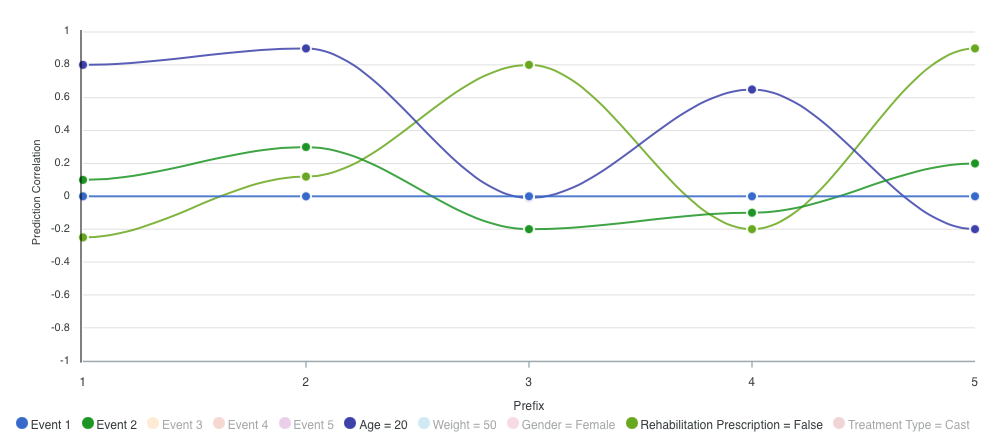}
        \caption{{\small Filtered view retaining only features Event 1, Event 2, Age, and Rehabilitation Prescription.}}    
        \label{fig:appendix:filterModeD}
    \end{subfigure}
    \caption{\small Example views before and after filtering some features.} 
    \label{fig:appendix:filterMode}
\end{figure*}

Questions asked by the investigator (Q) and projected answers based on the information provided (A):
\begin{question_answer}
    \item How do you interpret the plot? 
    \itema When the risk assessment is required, the requests tend to be executed and not sent to credit collection; when the risk assessment is not executed, the requests tend to be sent to credit collection.
    \item Based on the information presented in the plot, what course of action would you suggest?
    \item Would you recommend to change the process to include mandatory risk assessment?
    \itema Yes, because cases that executed the risk assessment usually are not sent to credit collection.
\end{question_answer}

\subsection{Interaction Capabilities of the Interface}
The plots of our study were provided with three interaction capabilities: (i) Detailed view, (ii) Focused view, and (iii) Information Filter.

\paragraph{Detailed view}
\label{sec:appendix:detailed_view}
When the user hovers over an element in the plot the exact values of that element are displayed, see Figure~\ref{fig:appendix:hover}. This allows the user to quickly retrieve specific information from the plots.

\paragraph{Focused view}
Hovering does not only show additional information regarding a particular point (see Section~\ref{sec:appendix:detailed_view}). When hovering is done on the legend, it also leads to the element to be highlighted. Figure~\ref{fig:appendix:focusMode} shows an example where the line for Event 2 is highlighted while the others are not.

\paragraph{Information Filter}
\label{sec:appendix:information_filter}
When the user clicks the labels in the legend they are temporarily removed from the plot. This allows the user to lower the amount of information displayed. This interaction capability is particularly useful for Plot P2 (Figure~\ref{fig:appendix:filterMode}). Figures~\ref{fig:appendix:filterModeA} and~\ref{fig:appendix:filterModeC} show two examples of unfiltered views.  Figures~\ref{fig:appendix:filterModeB} and~\ref{fig:appendix:filterModeD} show the corresponding filtered views.

\setstretch{2.0}
\bibliographystyle{apacite}
\bibliography{references}

\end{document}